\documentclass[a4paper]{article}
\usepackage{amsmath,amsthm,amsfonts,thmtools}
\usepackage{amssymb}
\usepackage{amsmath, amsfonts}
\usepackage{amssymb}
\usepackage{float}
\usepackage[cmintegrals]{newtxmath}
\usepackage{mathrsfs}   
\usepackage{amsfonts}  
\usepackage{graphicx}
\usepackage[table]{xcolor} 
\usepackage{diagbox} 
\usepackage{listings}
\usepackage[font=footnotesize]{caption}
\usepackage[font=scriptsize]{subcaption}
\usepackage{euscript}
\usepackage{units}
\usepackage{enumerate}
\usepackage{enumitem}
\usepackage{breqn}
\usepackage{xcolor}
\usepackage[T1]{fontenc}
\usepackage{newtxmath}
\usepackage{todonotes}
\usepackage{booktabs}
\usepackage{xcolor} 
\usepackage{tikz}
\usepackage{comment}
\DeclareMathAlphabet{\mathscr}{OMS}{rsfs}{m}{n}
 
\DeclareMathAlphabet{\pazocal}{OMS}{zplm}{m}{n}
\usepackage[bookmarks=true,hidelinks]{hyperref}
\usepackage{bbm}
\usepackage{multirow}
\usepackage{algpseudocode}  
\usepackage{mathtools}
\usepackage{subcaption} 
\usepackage{graphicx}
\usepackage{pgfplots}
\usepackage{float}
\usepackage{algpseudocode}
\usepackage{bbold}
\usepackage{cite}
\usepackage{microtype}
\usepackage{rotating}
\usepackage{enumitem}
\usepackage{bm}
\usepackage{silence}
\usepackage{hyperref}
\providecommand{\keywords}[1]
{
  \small	
  \textbf{\textit{Keywords:}} #1
}
\numberwithin{equation}{section}
\newtheorem{theorem}{Theorem}[section]

\newtheorem{definition}[theorem]{Definition}
 
\numberwithin{equation}{section}

\newtheoremstyle{appendixstyle} 
  {3pt}   
  {3pt}   
  {\normalfont} 
  {}      
  {\bfseries} 
  {.}     
  {.5em}  
  {\thmnumber{#2}} 
\theoremstyle{appendixstyle}

\theoremstyle{definition}

\newtheoremstyle{myremarkstyle}{}{}{}{}{\bfseries}{.}{ }{}
\theoremstyle{myremarkstyle}
\declaretheorem[name=Remark,qed=$\blacksquare$,numberlike=theorem]{remark}

\makeatletter
\newcommand*{\intavg}{%
  \mint@l{-}{}%
}
\newcommand*{\mint@l}[2]{%
  \@ifnextchar\limits{%
    \mint@l{#1}%
  }{%
    \@ifnextchar\nolimits{%
      \mint@l{#1}%
    }{%
      \@ifnextchar\displaylimits{%
        \mint@l{#1}%
      }{%
        \mint@s{#2}{#1}%
      }%
    }%
  }%
}
\newcommand*{\mint@s}[2]{%
  \@ifnextchar_{%
    \mint@sub{#1}{#2}%
  }{%
    \@ifnextchar^{%
      \mint@sup{#1}{#2}%
    }{%
      \mint@{#1}{#2}{}{}%
    }%
  }%
}
\def\mint@sub#1#2_#3{%
  \@ifnextchar^{%
    \mint@sub@sup{#1}{#2}{#3}%
  }{%
    \mint@{#1}{#2}{#3}{}%
  }%
}
\def\mint@sup#1#2^#3{%
  \@ifnextchar_{%
    \mint@sub@sup{#1}{#2}{#3}%
  }{%
    \mint@{#1}{#2}{}{#3}%
  }%
}
\def\mint@sub@sup#1#2#3^#4{%
  \mint@{#1}{#2}{#3}{#4}%
}
\def\mint@sup@sub#1#2#3_#4{%
  \mint@{#1}{#2}{#4}{#3}%
}
\newcommand*{\mint@}[4]{%
  \mathop{}%
  \mkern-\thinmuskip
  \mathchoice{%
    \mint@@{#1}{#2}{#3}{#4}%
        \displaystyle\textstyle\scriptstyle
  }{%
    \mint@@{#1}{#2}{#3}{#4}%
        \textstyle\scriptstyle\scriptstyle
  }{%
    \mint@@{#1}{#2}{#3}{#4}%
        \scriptstyle\scriptscriptstyle\scriptscriptstyle
  }{%
    \mint@@{#1}{#2}{#3}{#4}%
        \scriptscriptstyle\scriptscriptstyle\scriptscriptstyle
  }%
  \mkern-\thinmuskip
  \int#1%
  \ifx\\#3\\\else_{#3}\fi
  \ifx\\#4\\\else^{#4}\fi  
}
\newcommand*{\mint@@}[7]{%
  \begingroup
    \sbox0{$#5\int\m@th$}%
    \sbox2{$#5\int_{}\m@th$}%
    \dimen2=\wd0 %
    \let\mint@limits=#1\relax
    \ifx\mint@limits\relax
      \sbox4{$#5\int_{\kern1sp}^{\kern1sp}\m@th$}%
      \ifdim\wd4>\wd2 %
        \let\mint@limits=\nolimits
      \else
        \let\mint@limits=\limits
      \fi
    \fi
    \ifx\mint@limits\displaylimits
      \ifx#5\displaystyle
        \let\mint@limits=\limits
      \fi
    \fi
    \ifx\mint@limits\limits
      \sbox0{$#7#3\m@th$}%
      \sbox2{$#7#4\m@th$}%
      \ifdim\wd0>\dimen2 %
        \dimen2=\wd0 %
      \fi
      \ifdim\wd2>\dimen2 %
        \dimen2=\wd2 %
      \fi
    \fi
    \rlap{%
      $#5%
        \vcenter{%
          \hbox to\dimen2{%
            \hss
            $#6{#2}\m@th$%
            \hss
          }%
        }%
      $%
    }%
  \endgroup
}

\def\XXint#1#2#3{{\setbox0=\hbox{$#1{#2#3}{\int}$ }
		\vcenter{\hbox{$#2#3$ }}\kern-.6\wd0}}

\usepackage{silence}

\renewcommand{\epsilon}{\varepsilon}
\renewcommand{\phi}{\varphi}

\pgfplotsset{compat=newest}
\usepackage{silence}

\begin{document}

\title{A Family of Adaptive Activation Functions for Mitigating Failure Modes in Physics-Informed Neural Networks}


\author{Krishna Murari}
\date{\today}

\maketitle
\medskip

\centerline{$\dagger$Department of Mathematics, IIT Madras, Chennai 600036, India}

\centerline{Email:kmurari2712@gmail.com}

\begin{abstract}
\noindent
Physics-Informed Neural Networks(PINNs) are a powerful and flexible learning framework that has gained significant attention in recent years. It has demonstrated strong performance across a wide range of scientific and engineering problems. In parallel, wavelets have been extensively used as efficient computational tools due to their strong approximation capabilities. Motivated by the common failure modes observed in standard PINNs, this work introduces a novel family of adaptive wavelet-based activation functions. The proposed activation functions significantly improve training stability and expressive power by combining trainable wavelet functions with either trainable or fixed hyperbolic tangent and softplus functions. Five distinct activation functions are developed within the PINN framework and systematically evaluated across four representative classes of partial differential equations (PDEs). Comprehensive comparisons using bar plots demonstrate improved robustness and accuracy compared to traditional activation functions. Furthermore, the proposed approach is validated through direct comparisons with baseline PINNs, transformer-based architectures such as PINNsFormer, and other deep learning models, highlighting its effectiveness and generality.
\end{abstract}

\medskip

\keywords{Deep learning, Adaptive Activation Functions, Wavelets, Softplus, PDEs.}

\section{Introduction}\label{sec:intro}
\noindent
Numerical solutions of PDEs have long been a central topic in both science and engineering. Owing to the complexity of the underlying mesh, classical discretization techniques, such as the finite element method (FEM)~\cite{bathe2007finite} and pseudo-spectral methods~\cite{fornberg1998practical}, often become computationally expensive in high-dimensional settings. The Machine Learning (ML) technique has been adopted in recent years for resolving these limitations in scientific computing. The fundamental publications in this regard are \cite{lagaris1998artificial}, \cite{han2018solving}, and \cite{raissi2019physics}. Among these techniques, PINNs~\cite{raissi2019physics} leverage advances in scientific machine learning to develop mesh-independent solution strategies for traditional numerical methods~\cite {Hermeline2000}, enabling applications to high-dimensional PDEs and irregular geometries. To approximate PDE solutions, PINNs directly incorporate governing physical laws into neural network(NN) training. Different training approaches for PINNs were studied in various previous studies~\cite{mao2020physics}, ~\cite{wang2021understanding}, and ~\cite{wang2022and}, wherein improved optimization patterns can be found during training. This ability of PINNs to handle direct and inverse problems makes the PINN technique highly successful across many applications in the physical sciences and engineering. Therefore, PINNs and its various versions can be found helpful in solving multiple real-time applications involving radiation transfer problems~\cite{mishra2021physics}, medical imaging~\cite{zhang2025personalized}, quantum systems~\cite{yang2020physics}, and many more complex physical phenomena. Several enhanced PINN variants have been proposed, including  XPINNs\cite{jagtap2020extended}, conservative PINNs \cite{jagtap2020conservative}, and ML-PINN \cite{gao2025ml}. Architectural modifications have been introduced to improve the expressiveness and robustness of PINN-based models~\cite{wang2025mixed}. In these frameworks, activation functions play an essential role in expressive capability, gradient flow, and convergence properties. A substantial body of work has examined scalable, adaptive, and other activation formulations, with notable contributions summarized in~\cite{Wang2025, szandala2020review, Dushkoff2016, Parisi2024, Srivastava2014, dugas2000incorporating, biswas2020tanhsoft, Jagtap2020rsp, Wang2023, Li2013, Zafar2025, Uddin2023, Wang2021, Zhao2024PINNsFormer, wang2025mixed, kiliccarslan2024parametric, apicella2021survey, farea2025learnable}. Significant theoretical developments have also been made, focusing on understanding the mathematics underlying PINN convergence. For instance, a manuscript like \cite{de2024error} deals with advancing an estimate of error for the flow problem described by the Navier-Stokes equations, while another one like \cite{DeRyck2024} would advance a numerical study of PINNs at large. 
Nevertheless, some previous investigations suggest possible shortcomings of PINNs when confronted with challenges involving oscillatory patterns, high-frequency features, and multi-scale aspects~\cite{fuks2020limitations, raissi2018deep, mcclenny2020self, krishnapriyan2021characterizing, wang2022and}. As such, the simplicity of the exact solution warrants more specific details within the obtained forecast. Currently available advances of PINNs are primarily founded on the application of multilayer perceptrons (MLPs). These have proven to be successful tools, though to some extent, due to possible difficulties in the network's ability to discern particular characteristics in more complicated scenarios. \cite{krishnapriyan2021characterizing} discovered that physics-based regularization may produce optimization that poses a significant optimization issue, which reduces the optimization performance in a learning scenario. The gradients associated with each of the loss function elements become extremely uneven; hence, optimization might be challenging when applying the standard gradient-based optimization process, approaches such as the FEM inherently account for such a scenario, as the global solution at each point evolves sequentially, where the solution at $(t+\Delta t)$ depends on the solution at $(t)$ in the same system. In contrast, a standard PINN uses an MLPs, which operates point-to-point and inherently lacks temporal information within PDEs. This limitation affects the global spread of initial condition information, and as a consequence, PINNs are known to perform well in the initial region but suffer accuracy issues over time, producing smooth or incorrect predictions. These are mainly issues pertaining to the formulation of data representations and the design of the regularization loss functions. For the resolution of these aforesaid issues, various techniques were used, as presented in the studies included in the literature such as~\cite{han2018solving, lou2021physics, wang2021understanding, wang2022and, wang2022auto}. The \textit{Seq2Seq} model, as presented by the study~\cite{krishnapriyan2021characterizing}, is based on the sequential multi-network, which adds complexity to the model, leading to an accumulation of possibilities for error. Similarly, the \textit{Neural Tangent Kernel (NTK)} approach \cite{wang2022and} constructs a kernel matrix $K \in \mathbb{R}^{D \times P}$, where $D$ represents the sample size and $P$ the number of model parameters, leading to significant scalability challenges as either dimension grows. Therefore, scalability issues arise as either of these two quantities increases. Various models developed based on the Transformer paradigm, follow a physics-constrained \cite{Yang2022PhysicsConstrainedDynamics} approach to the design of dynamical learning models. Moreover, their method is also compared to the performance achieved by PINNs~\cite{raissi2019physics}, First Layer Sine Method(FLS)~\cite{wong2022learning}, and Quadratic Residual Networks(QRes)~\cite{bu2021quadratic} in \cite{Zhao2024PINNsFormer}. Moreover, computationally efficient architectures such as PINN-Mamba~\cite{Xu2025SubSequential} and ML-PINN~\cite{gao2025ml} provide a practical approach to addressing the challenges posed by large-scale PDE solving. 

A crucial ingredient in modern scientific computing is wavelets, which are used to decompose signals into components at different scales. A wavelet basis can be generated with the help of two parent wavelets. One wavelet is the physical or time domain representation, whereas the other wavelet serves as the basis or the scale representation. Because both wavelets are localized in both the physical and scale domains, they offer efficient analysis tools in numerical analysis, signal processing, and applied mathematics. Over the years, many classical contributions have shaped the development of wavelet theory, starting from the early work of Haar~\cite{haar1909} and later by Grossmann and Morlet~\cite{grossmann1984decomposition}, Meyer~\cite{meyer1989wavelets}, Mallat~\cite{mallat2002theory}, and Daubechies~\cite{daubechies1992ten}. Wavelets are widely used in ML techniques. For example, GaborPINN~\cite{huang2023gaborpinn} leverages multiplicative filtered networks to improve computational efficiency. Furthermore, \cite{Uddin2023} and \cite{Zhao2024PINNsFormer} developed wavelet-based activation functions, which further manifested the potential of wavelet-driven models. Moreover, ~\cite{tripura2023wavelet} proposed a wavelet neural operator to address parameterized PDEs arising in mechanics. In this work, we make use of several commonly employed wavelet functions, namely the \emph{Morlet wavelet}, the \emph{Mexican hat wavelet}, the \emph{Gaussian wavelet}, the \emph{Hermite wavelet}, and the \emph{Gabor wavelet}, due to their favorable mathematical characteristics and practical relevance.\\

\noindent
\textbf{The key contribution of this work is as follows :}

 \textbf{1. Design of adaptive custom activation functions:}
 In this work, we have constructed five novel custom activation functions. These functions are constructed by combining trainable wavelets or wavelet-inspired functions with the hyperbolic tangent function. The wavelet and wavelet-inspired functions employed in this study include the Mexican hat, Morlet, Hermite, Gabor, and Gaussian functions, which are described in detail in Section~\ref{sec:5}. Additionally, the softplus function is incorporated to ensure that the trainable parameter remains strictly positive. The newly developed activation functions in this work are denoted as \texttt{SoftMexTanh}, \texttt{SoftMorTanh}, \texttt{SoftGassTanh}, \texttt{SoftGaborTanh}, and \texttt{SoftHermTanh}. Their ablation variants, in which the hyperbolic tangent component is fixed (non trainable), are referred to as \texttt{SoftMexTanhW}, \texttt{SoftMorTanhW}, \texttt{SoftGassTanhW}, \texttt{SoftGaborTanhW}, and \texttt{SoftHermTanhW}. In this work, we treat the terms \textcolor{blue}{learnable, trainable, and adaptive} as equivalent and use them interchangeably.

\textbf{2. Mitigating failure modes of PINNs in solving PDEs:} PINNs with conventional activation functions such as $\tanh$ have been found to exhibit failure modes of PINNs \cite{krishnapriyan2021characterizing, Zhao2024PINNsFormer}. We employ newly developed activation functions in PINNs to mitigate known failure modes. Using these activation functions, we successfully simulate four representative PDEs, including the reaction, wave, convection, and Navier–Stokes equations, thereby addressing these failure modes.

\textbf{3. Robustness and validation:} Our proposed activation functions for PINNs are compared with other deep-learning-based methods, including standard PINNs~\cite{raissi2019physics} and PINNsFormer~\cite{Zhao2024PINNsFormer}. In addition, the proposed results outperform QRes~\cite{bu2021quadratic} and FLS~\cite{wong2022learning}, which exhibit performance comparable to standard PINNs (see~\cite{Zhao2024PINNsFormer}). Apart from these approaches, the proposed methodology yields improved results relative to PINN-Mamba~\cite{Xu2025SubSequential} and ML-PINN~\cite{gao2025ml}, thereby establishing its robustness and efficiency(see~\cite{gao2025ml}). Overall, our approach yields more accurate results than the baseline methods. Loss and error statistics for different activation functions are examined and summarized using bar plots.\\

\noindent
\textbf{The structure of this work is as follows:} Section~\ref{sec:2} summarizes existing literature, and Section~\ref{sec:3} introduces the PDE formulations together with their mathematical representations.
Section~\ref{sec:4} describes the PINNs formulation and the overall solution methodology. 
Section~\ref{sec:5} discusses the network architecture, model design, and the proposed activation functions. 
Section~\ref{sec:6} presents the numerical experiments and analyzes the corresponding results, including a comparative study of different activation functions illustrated through bar plots.
Section~\ref{sec:7} outlines the key outcomes and delivers the concluding discussion.

\section{Related work}\label{sec:2}
However, the performance of PINNs heavily depends on several essential factors, including NN architecture, weight initialization procedures, optimization techniques, and the activation function. This activation function is the vital core of the NNs, where non-linearity takes its primary importance in relation to gradient propagation~\cite{szandala2020review}, the speed of integration~\cite{Parisi2024}, and the ability of the NNs~\cite{Srivastava2014} to express the data most desirably. Sigmoid and tanh activation functions frequently suffer from vanishing gradients ~\cite{Srivastava2014}, unlike ReLU activation functions.
The softplus activation function and other variants of this function have been introduced in \cite{dugas2000incorporating} and further in \cite{biswas2020tanhsoft}. An extensive survey of adaptive activation mechanisms is discussed in~\cite{apicella2021survey}. Research on adapting activation functions to improve NN performance across various problems has been discussed in several papers~\cite{Jagtap2020rsp, Wang2023, Li2013}. Research carried out in~\cite{Uddin2023} has demonstrated that using wavelet functions in PINNs is effective in solving. Finally, the work in ~\cite{Zafar2025} presents various hybrid activation functions that combine the desired attributes of standard activations, giving better convergence characteristics and lower residual loss for PINNs. Standard activating functions have been tested in applications such as the Swift–Hohenberg and Burgers equations.  \cite{kiliccarslan2024parametric} proposed Parametric RSigELU, a novel trainable activation scheme. Learnable activation functions \cite{farea2025learnable} have been introduced in PINNs to enhance convergence and accuracy when solving PDEs.
This formulation is motivated by the real Fourier transform, which states that any signal can be represented as an integral combination of sine and cosine components at different frequencies, thereby enabling the wavelet activation to approximate arbitrary functions. Besides PINNsFormer, other PINNs-based architectures like the attention-based model~\cite{srati2025computational}, PINN-Mamba~\cite{Xu2025SubSequential}, and ML-PINN~\cite{gao2025ml} also adopt this wavelet activation. Motivated by this recent development, in this work, we propose a family of efficient adaptive activation functions, wavelet-based or inspired, combined with softplus and hyperbolic tangent variants to mitigate some modes of failure of PINNs when solving PDEs, enhancing their robustness to initialization and convergence behavior at the same time.

\section{The models: PDEs setup}\label{sec:3}
The 1D reaction, 1D wave, 1D convection, and 2D Navier–Stokes equations are considered as benchmark problems in this work. These equations are selected because standard PINNs have been reported to fail in solving such equations with high accuracy in~\cite{krishnapriyan2021characterizing}, hence serving as ideal test cases for investigating the failure modes of PINNs. Below are selected test cases used to examine the accuracy, convergence properties, and robustness of PINNs with proposed adaptive wavelet-based activation functions against known training difficulties. The models are listed below \cite{krishnapriyan2021characterizing, Zhao2024PINNsFormer, gao2025ml}:

\noindent
\textbf{1D Reaction Equation:}
The reaction equation constitutes a nonlinear hyperbolic-type PDE arising in the mathematical description of reaction phenomena. When supplemented with periodic boundary conditions(\texttt{BC}), it takes the form
\begin{equation}\label{eqn:1dreactio}
\begin{aligned}
    \frac{\partial u}{\partial t} - \rho\, u(1-u) &= 0, 
    && \forall\, x \in [0,2\pi],\; t \in [0,1], \\
  \texttt{IC:}\quad 
   u(x,0) &= h(x), \\
  \texttt{BC:}\quad 
   u(0,t) &= u(2\pi,t)
\end{aligned}
\end{equation}
Here, $\rho$ represents the reaction parameter and is fixed to $5$ in all numerical experiments. The corresponding analytical solution is
\begin{equation}
    u_{\texttt{analytical}}(x,t)
    = \frac{h(x)\, \exp\!\left( \rho t\right)}{h(x)\, \exp\!\left( \rho t\right) + 1 - h(x)},
\end{equation}
where $h(x)$ specifies the initial condition(\texttt{IC}), expressed as
$\exp\!\left(-\frac{(x-\pi)^2}{2(\pi/4)^2}\right)$.

\noindent
\textbf{1D Wave Equation:}
The 1D wave equation is a prototypical hyperbolic PDE modeling wave motion in one spatial dimension. It is encountered in many applications, including acoustics, electro magnetics, and seismology. Under periodic boundary conditions, the equation takes the form
\begin{equation}\label{eqn:1dwave}
\begin{aligned}
    \frac{\partial^{2} u}{\partial t^{2}}
    - 4\,\frac{\partial^{2} u}{\partial x^{2}} &= 0,
    && \forall\, x \in [0,1],\; t \in [0,1], \\
    \texttt{IC:}\quad 
    u(x,0) &= \sin(\pi x) + \tfrac{1}{2}\sin(\beta'\pi x), \\
    \frac{\partial u(x,0)}{\partial t} &= 0, \\
    \texttt{BC:}\quad 
    u(0,t) &= u(1,t) = 0.
\end{aligned}
\end{equation}
Here, $\beta'$ denotes the wave frequency parameter and in our experiments we 
choose $\beta' = 3$. The corresponding analytical solution is
\begin{equation}
    u(x,t)
    = \sin(\pi x)\cos(2\pi t)
      + \frac{1}{2}\sin(\beta'\pi x)\cos(2\beta'\pi t).
\end{equation}
\noindent
\textbf{1D Convection Equation:}\label{eqn:con}
The 1D convection equation belongs to the class of hyperbolic PDEs. Such equations commonly arise in simplified transport-type modeling frameworks. The problem is considered with periodic boundaries and is formulated as
\begin{equation}
\begin{aligned}
    \frac{\partial u}{\partial t} 
    + \beta' \frac{\partial u}{\partial x} &= 0,
    && \forall\, x \in [0, 2\pi],\; t \in [0,1], \\
    \texttt{IC:}\quad u(x,0) &= \sin(x), \\
    \texttt{BC:}\quad u(0,t) &= u(2\pi,t).
\end{aligned}
\end{equation}
Here, $\beta'$ denotes the convection coefficient. As $\beta'$ increases, the 
solution oscillates at a higher frequency, making the problem increasingly challenging for traditional PINNs architectures to approximate. The convection equation admits an analytical solution expressed as
\[
    u(x,t) = \sin\!\bigl(x - \beta' t\bigr),
\]
where we set \(\beta' = 50\) throughout all numerical experiments.\\
\noindent
\textbf{2D Navier-Stokes Equations:}
The equations constitute a system of nonlinear PDEs governing incompressible fluid flow in two spatial dimensions. As fundamental components of fluid dynamics, such equations provide critical models for describing the behavior of air, water, and other fluid systems, enabling analysis and simulation in numerous scientific and engineering applications. The governing equations are
\begin{equation}
\begin{aligned}\label{eqn:ns}
    \frac{\partial u}{\partial t}
    + \lambda_{1}\!\left(u\frac{\partial u}{\partial x}
    + v\frac{\partial u}{\partial y}\right)
    &= -\frac{\partial p}{\partial x}
    + \lambda_{2}\!\left(\frac{\partial^{2}u}{\partial x^{2}}
    + \frac{\partial^{2}u}{\partial y^{2}}\right), \\[6pt]
    \frac{\partial v}{\partial t}
    + \lambda_{1}\!\left(u\frac{\partial v}{\partial x}
    + v\frac{\partial v}{\partial y}\right)
    &= -\frac{\partial p}{\partial y}
    + \lambda_{2}\!\left(\frac{\partial^{2}v}{\partial x^{2}}
    + \frac{\partial^{2}v}{\partial y^{2}}\right).
\end{aligned}
\end{equation}
In this formulation, $u(t,x,y)$ and $v(t,x,y)$ correspond to the velocity components in the $x$- and $y$-directions, whereas $p(t,x,y)$ represents the pressure field. Throughout the experiments, the parameters are fixed at $\lambda_{1}=1$ and $\lambda_{2}=0.01$. A closed-form analytical solution is unavailable for this system. The reference (ground-truth) solution used for evaluation is taken from the numerical simulation provided in~\cite{raissi2019physics, Zhao2024PINNsFormer}.

\section{PINNs approximation: An overview of an approach to solve PDEs}\label{sec:4}
\label{sec:method}
\begin{figure}[htbp]
\centering
\includegraphics[height=0.47\textheight]{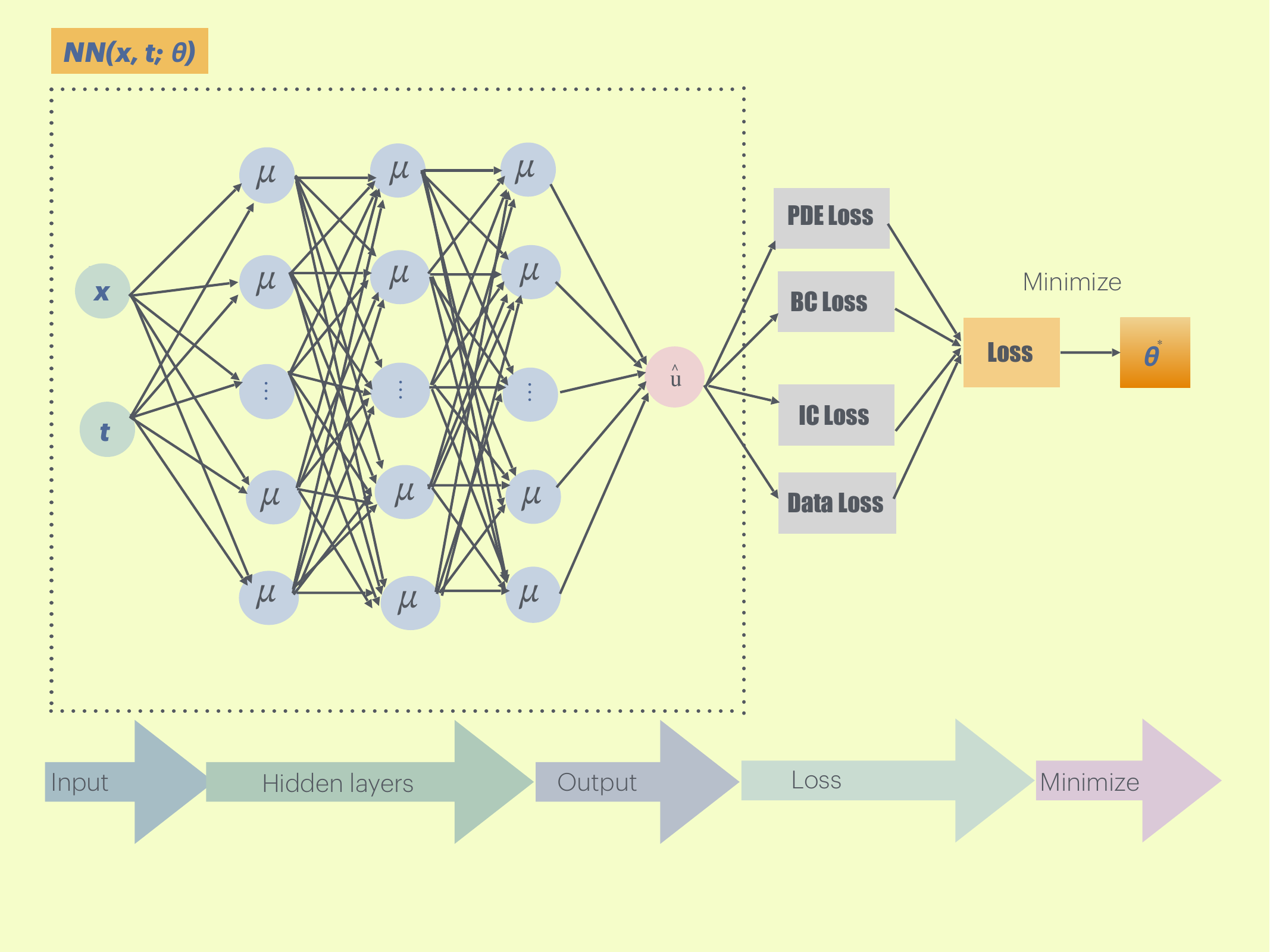}
\caption{Illustration of the PINNs framework.}
\label{fig:figttt}
\end{figure}

Let \(\mathrm{D} \subset \mathbb{R}^{d}\) be an open and bounded spatial domain with 
boundary \(\partial \mathrm{D}\). A general time-dependent PDE can be written as
\begin{equation}
\begin{aligned}
\mathbf{D}\!\left[u(x,t)\right] &= f(x,t),
&\qquad& (x,t) \in \mathrm{D}, \\[6pt]
\mathbf{B}\!\left[u(x,t)\right] &= g(x,t),
&\qquad& (x,t) \in \partial \mathrm{D}, \\[6pt]
\mathbf{I}\!\left[u(x,0)\right] &= h(x),
&\qquad& x \in \mathrm{D}.
\end{aligned}
\end{equation}
Here, \(u(x,t)\) denotes the unknown solution, \(\mathbf{D}\) represents the differential operator associated with the governing equations, \(\mathbf{B}\) enforces the \texttt{BC}, and \(\mathbf{I}\) enforces the \texttt{IC}.
The sets \(\{x_i,t_i\} \in \mathrm{D}\) correspond to the interior (residual) 
collocation points, while \(\{x_i,t_i\} \in \partial\mathrm{D}\) represent the 
boundary or initial condition points. Let \( \hat{u}(x;\theta) \) represent a feed-forward network with parameters \( \theta \) used to model the solution \( u(x) \). 
The input variable is \( x \in \mathbb{R}^{d_{\text{in}}} \) and the trainable network parameters are \( \theta = \{W_\ell, b_\ell\}_{\ell=1}^{L} \), where each \( W_\ell \) and \( b_\ell \) represents the weight matrix and bias vector of layer \( \ell \), respectively. The network output \( \hat{u}(x;\theta) \in \mathbb{R}^{d_{\text{out}}} \) is obtained by composing affine transformations with a nonlinear activation function \( \mu \). The network has depth \( L \), consisting of an input mapping \( \eta_0(x) = x \in \mathbb{R}^{d_{\text{in}}} \), followed by \( L-1 \) hidden layers and an output layer \( \eta_L(x) \in \mathbb{R}^{d_{\text{out}}} \). Each layer \( \ell \) has width \( n_\ell \), with 
\( W_\ell \in \mathbb{R}^{n_\ell \times n_{\ell-1}} \) and \( b_\ell \in \mathbb{R}^{n_\ell} \). The network is defined recursively as
\begin{equation}
\begin{aligned}
\eta_0(x) &= x,\\[6pt]
\eta_\ell(x)  &= \mu\!\bigl( W_\ell \eta_{\ell-1}(x) + b_\ell \bigr), 
\qquad
1 \le \ell \le L, \\[6pt]
\eta_L(x) &= \mu(W_L \eta_{L}(x) + b_L).
\end{aligned}
\end{equation}
Information flows from the input through the sequence of hidden layers and finally 
to the output layer, producing the approximation \( \hat{u} \). Let $\hat{u}(x;\theta)$ denote the NN approximation of the
solution, where $\theta$ represents the collection of trainable parameters
(weights and biases). Thus, the NN provides an approximation to the solution of the governing PDEs, expressed as
\[
\hat{u}(x;\theta) = \eta_L(x).
\]
 PINNs incorporate the physical constraints by minimizing a composite loss consisting of PDE residual, boundary, and initial condition terms. Backpropagation is employed during training to iteratively adjust the model parameters in accordance with a specified objective function. Unlike conventional NNs, a PINNs must additionally satisfy the physical constraints 
encoded in the governing PDEs. Figure \ref{fig:figttt} shows the schematic structure of the PINN. To enforce these constraints, the loss function 
is constructed using the residual of the PDEs together with the imposed 
\texttt{IC} and \texttt{BC}, all evaluated at selected collocation points. 
\begin{equation}\label{eq:loss1}
\mathscr{L}_{\text{PINNs}}
=
\lambda_R\mathscr{L}_{R}
+
\lambda_B\mathscr{L}_{B}
+
\lambda_I\mathscr{L}_{I}.
\end{equation}
Here, $\lambda_R$, $\lambda_B$, and $\lambda_I$ are non-negative weighting parameters that regulate the relative contributions of the PDE residual, boundary, and initial condition terms. The corresponding error components are quantified through mean-squared error (MSE) measures defined as

\begin{equation}
\begin{aligned}
\mathscr{L}_{R}&=\frac{1}{{N}_{R}}\sum_{i=1}^{N_{R}}\big\|\mathbf{D}\!\left[\hat{u}(x_i,t_i)\right]- f(x_i,t_i)\big\|^{2},\\
\mathscr{L}_{B} &=\frac{1}{N_{B}}\sum_{i=1}^{N_{B}}
\big\|\mathbf{B}\!\left[\hat{u}(x_i,t_i)\right]- g(x_i,t_i)\big\|^{2},\\
\mathscr{L}_{I}&=\frac{1}{N_{I}}\sum_{i=1}^{\mathbf{N}_{I}}\big\|\mathbf{I}\!\left[\hat{u}(x_i,0)\right]- h(x_i)\big\|^{2}.
\end{aligned}
\end{equation}
In this formulation, $N_R$, $N_B$, and $N_I$ indicate the respective number of residual, boundary, and initial points. For convenience, we set
\(\lambda_R = \lambda_B = \lambda_I = 1\) in all experiments. These parameters control the relative contribution of each loss component. The mapping $\hat{u}(x,t)$ receives the spatial–temporal coordinates as inputs and yields an approximation of the governing PDEs solution. Training the PINNs involves an optimization procedure over the model parameters with the objective of reducing the total loss $\mathscr{L}_{\text{PINNs}}$ defined in~\eqref{eq:loss1}.
\section{Architecture framework and models design in PINNs:}\label{sec:5}
The following section describes the PINNs framework employed in this work. Designing an effective PINNs is essential for achieving an accurate and stable approximation of the underlying PDEs solution. Network behavior is influenced by architectural and training choices, including layer depth, neuron count, and the placement of collocation points. Activation function selection is essential to ensure stable training and improved expressivity.
\subsection{Loss}
In this section, we discuss the loss structure for PDEs. Now residual structure of 1D reaction equations is 
\begin{equation}
\begin{aligned}
\mathscr{R}(x,t) \coloneqq \hat{u}_t(x,t) - \rho\,\hat{u}(x,t)\bigl(1-\hat{u}(x,t)\bigr),
\end{aligned}
\end{equation}
We can write,
\begin{equation}
\begin{aligned}
\mathscr{L}_{R}
\;=\;
\frac{1}{N_{R}}\sum_{j=1}^{N_{R}}
\bigl|
\mathscr{R}(x_j,t_j)
\bigr|^{2}
\;&=\;
\frac{1}{N_{R}}\sum_{j=1}^{N_{R}}
\Bigl|
u_t(x_j,t_j) - \rho\,u(x_j,t_j)\bigl(1-u(x_j,t_j)\bigr)
\Bigr|^{2},\\
\mathscr{L}_{B}
\;&=\;
\frac{1}{N_{B}}\sum_{j=1}^{N_{B}}
\Bigl(
u(0,t_j) - u(2\pi,t_j)
\Bigr)^{2},\\
\mathscr{L}_{I}
\;&=\;
\frac{1}{N_{I}}\sum_{j=1}^{N_{I}}
\Bigl(
u(x_j,0) - h(x_j)
\Bigr)^{2}.
\end{aligned}
\end{equation}
Now, we can write loss of 1D reaction equation as \ref{eq:loss1}
\begin{equation*}\label{eq:loss2}
\mathscr{L}_{\text{PINNs}}
=
\lambda_R\mathscr{L}_{R}
+
\lambda_B\mathscr{L}_{B}
+
\lambda_I\mathscr{L}_{I}.
\end{equation*}
In a similar manner, the loss functions for the wave, convection and Navier-Stokes equations can be defined.
\subsubsection{Activation functions, wavelets and wavelet-inspired functions}
\begin{figure}[htbp]
\centering
\begin{subfigure}[b]{0.45\textwidth}
    \includegraphics[width=\textwidth]{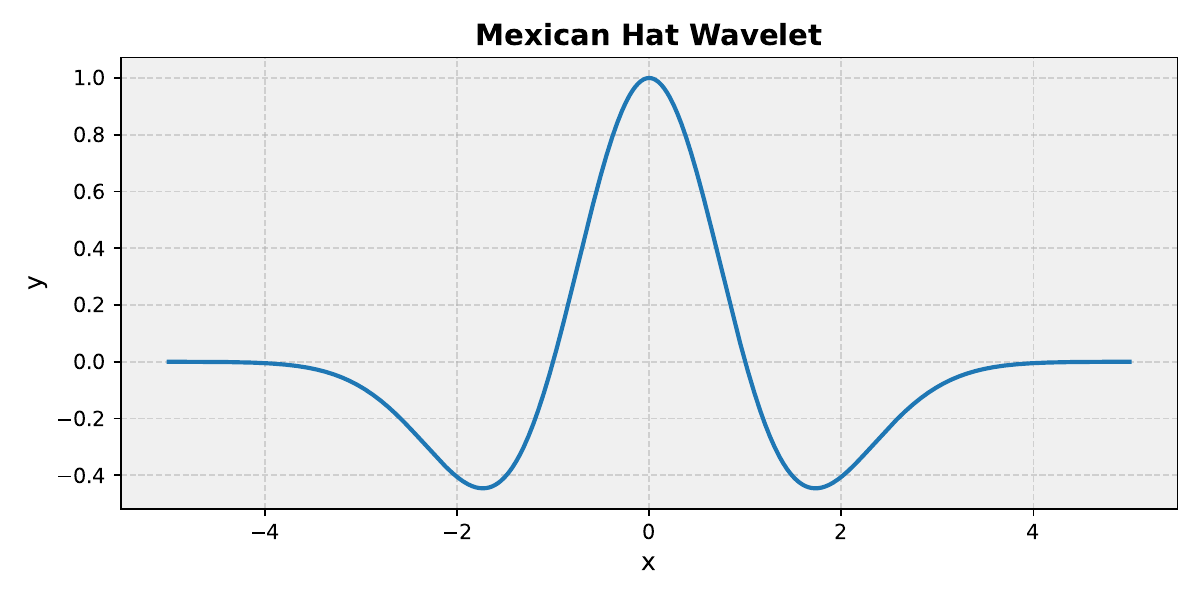}
    \caption{Mexican hat wavelet}
    \label{fig:mexican_hat}
\end{subfigure}
\hfill
\begin{subfigure}[b]{0.45\textwidth}
    \includegraphics[width=\textwidth]{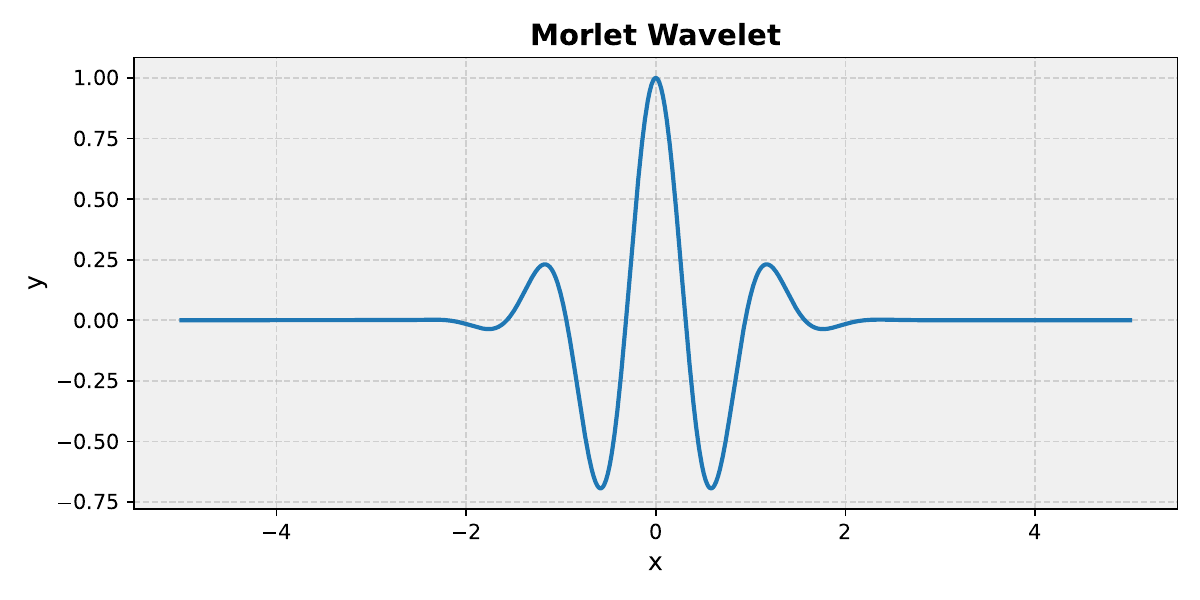}
    \caption{Morlet wavelet}
    \label{fig:morlet}
\end{subfigure}
\caption{Comparison between the Mexican hat and Morlet wavelets.}
\label{fig:mexican_morlet}
\end{figure}

\begin{figure}[htbp]
\centering

\begin{subfigure}[b]{0.45\textwidth}
    \includegraphics[width=\textwidth]{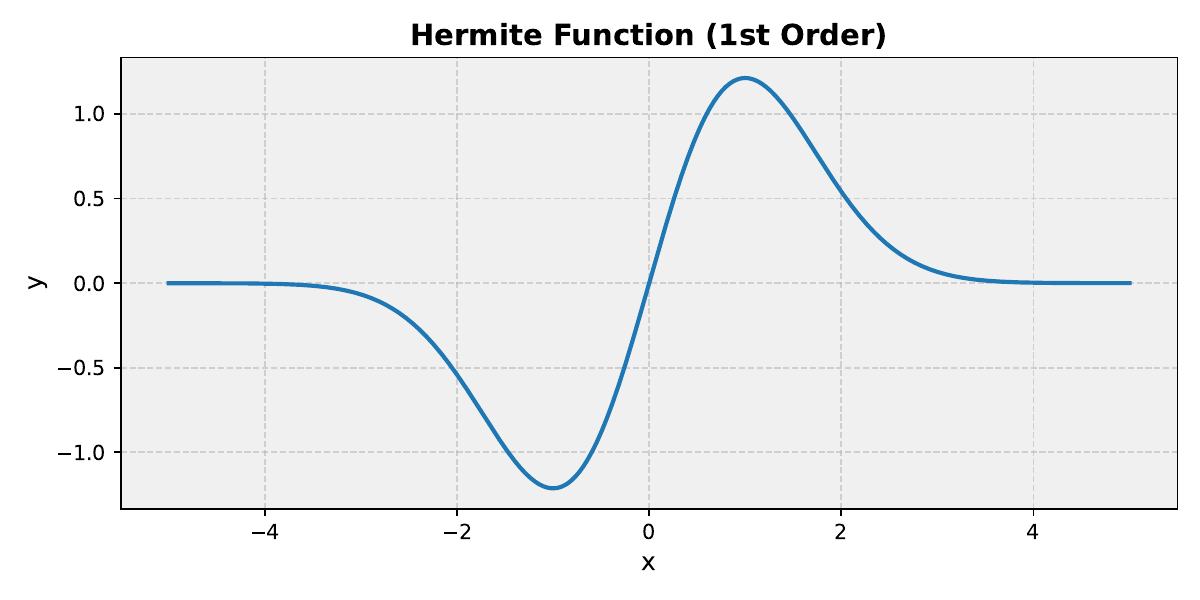}
    \caption{First-order Hermite function}
    \label{fig:hermite1}
\end{subfigure}
\hfill
\begin{subfigure}[b]{0.45\textwidth}
    \includegraphics[width=\textwidth]{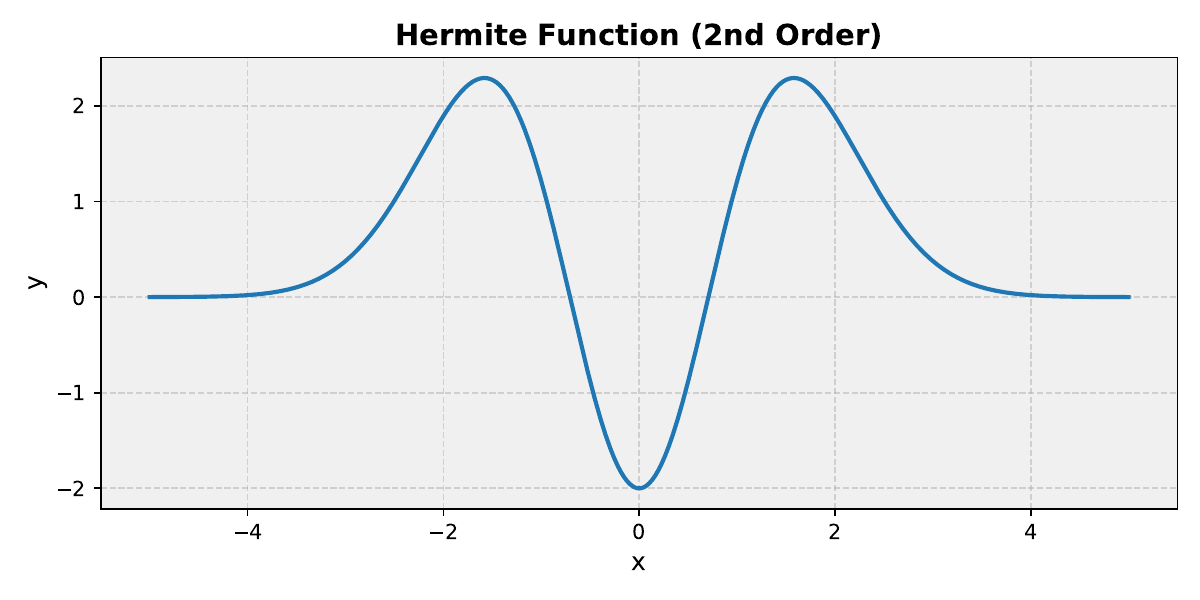}
    \caption{Second-order Hermite function}
    \label{fig:hermite2}
\end{subfigure}

\vspace{0.4cm}

\begin{subfigure}[b]{0.45\textwidth}
    \includegraphics[width=\textwidth]{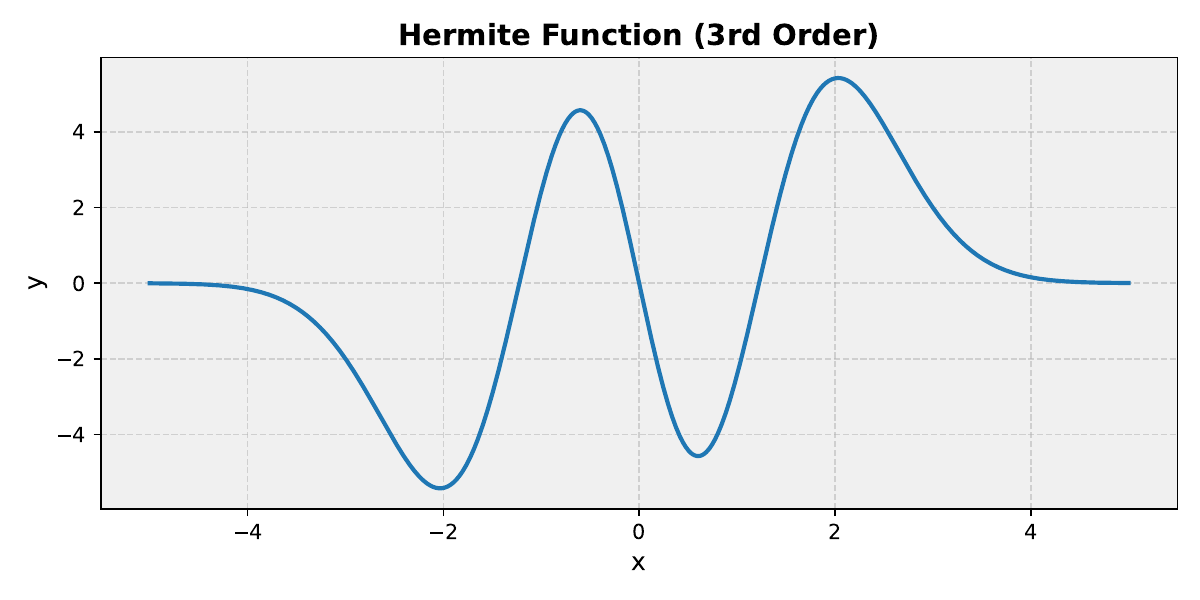}
    \caption{Third-order Hermite function}
    \label{fig:hermite3}
\end{subfigure}
\hfill
\begin{subfigure}[b]{0.45\textwidth}
    \includegraphics[width=\textwidth]{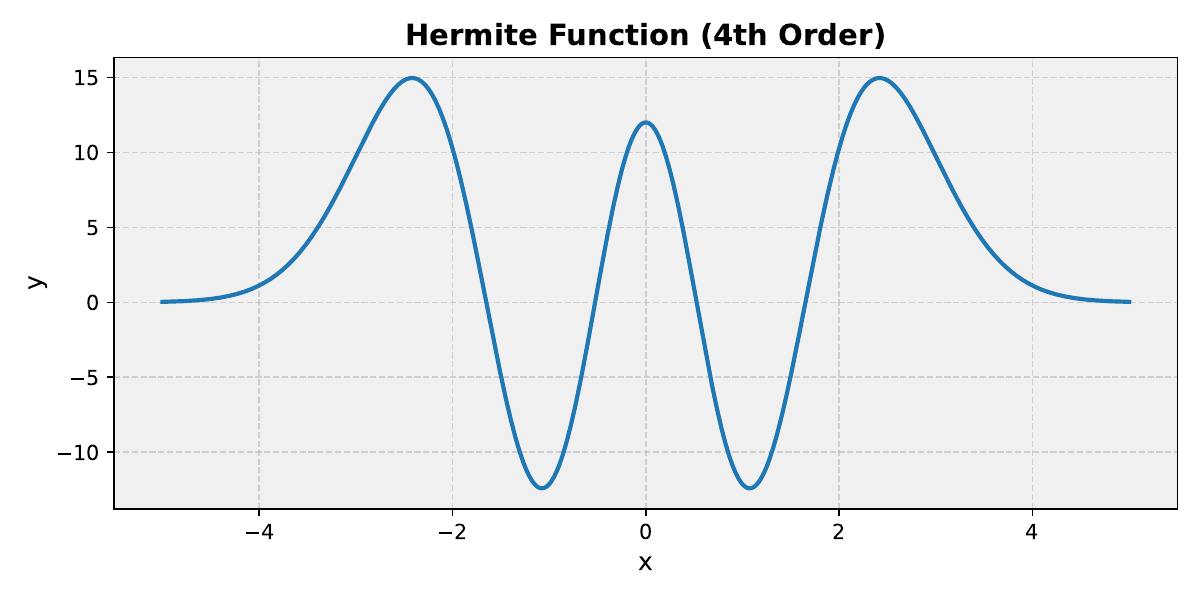}
    \caption{Fourth order Hermite function}
    \label{fig:hermite4}
\end{subfigure}
\caption{Hermite functions of orders 1, 2, 3, and 4.}
\label{fig:hermite_wavelets}
\end{figure}

\begin{figure}[htbp]
\centering
\begin{subfigure}[b]{0.45\textwidth}
    \includegraphics[width=\textwidth]{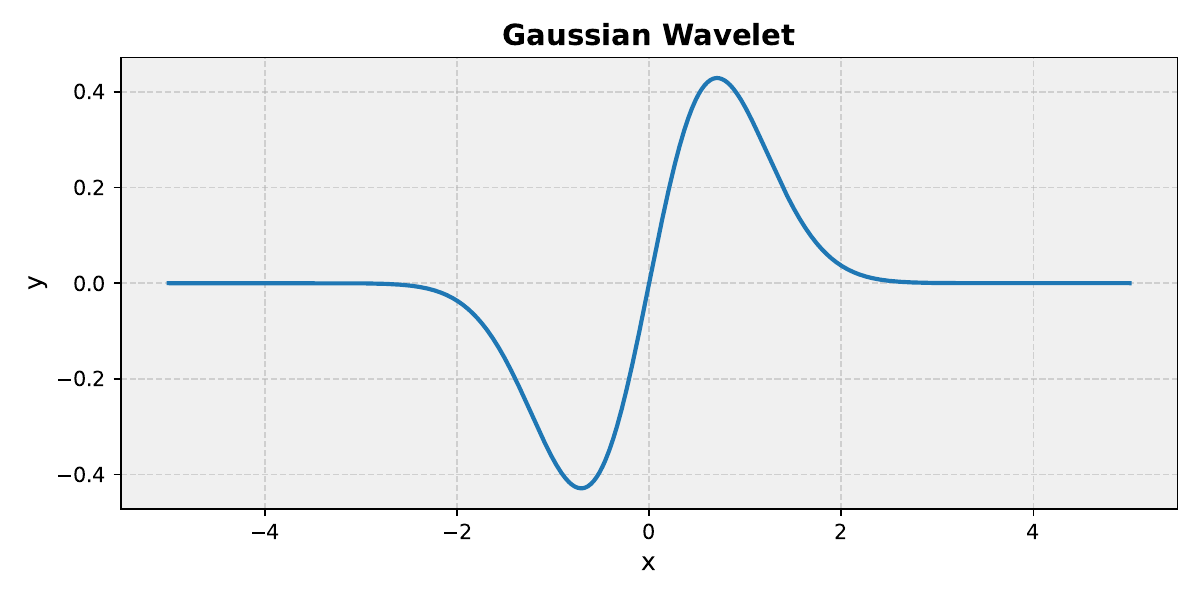}
    \caption{Gaussian wavelet}
    \label{fig:gaussian}
\end{subfigure}
\hfill
\begin{subfigure}[b]{0.45\textwidth}
    \includegraphics[width=\textwidth]{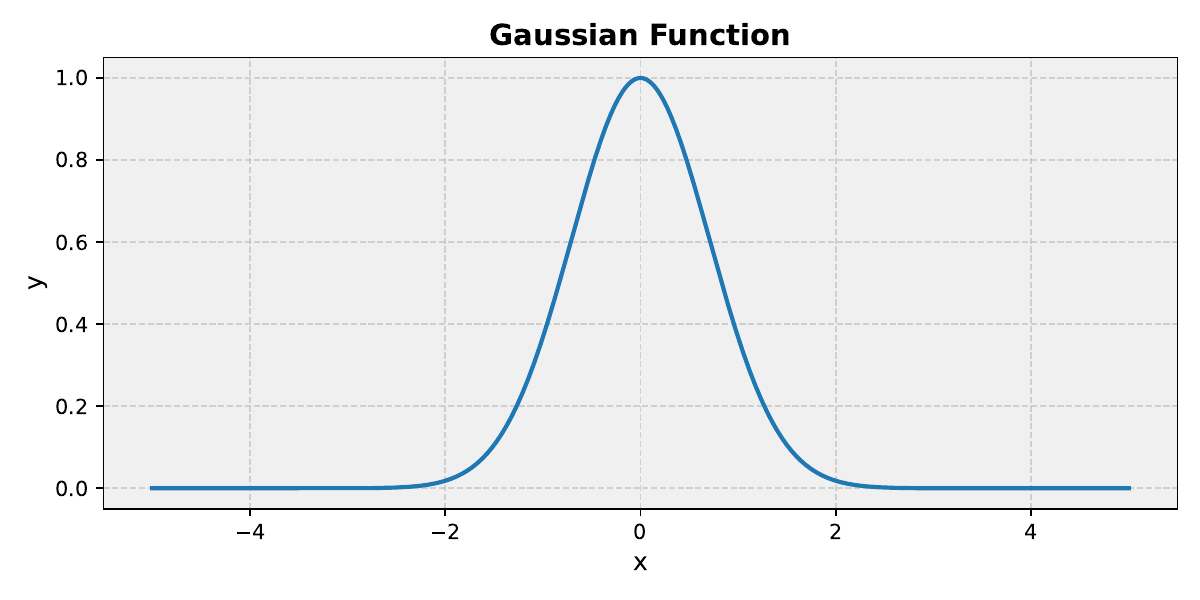}
    \caption{Gaussian function}
    \label{fig:gaussian_funtion}
\end{subfigure}

\caption{Comparison between the Gaussian wavelet and function}
\label{fig:gaussian_n}
\end{figure}

\begin{figure}[htbp]
\centering
\begin{subfigure}[b]{0.45\textwidth}
    \includegraphics[width=\textwidth]{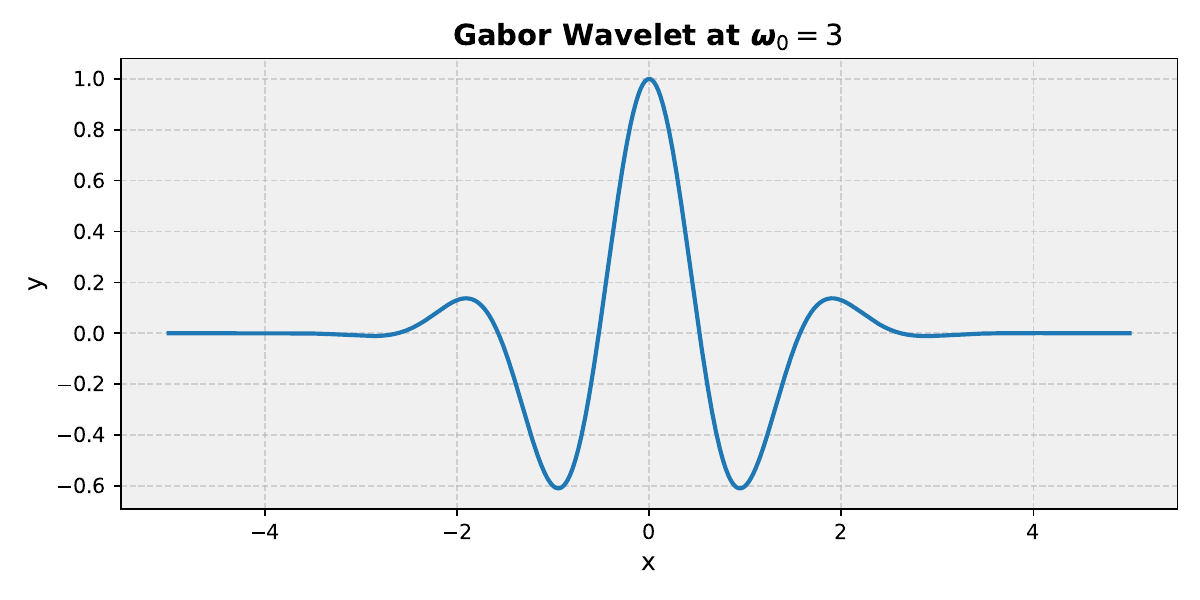}
    \caption{Gabor wavelet at $\omega_{0} =3$}
    \label{fig:gabor_5}
\end{subfigure}
\hfill
\begin{subfigure}[b]{0.45\textwidth}
    \includegraphics[width=\textwidth]{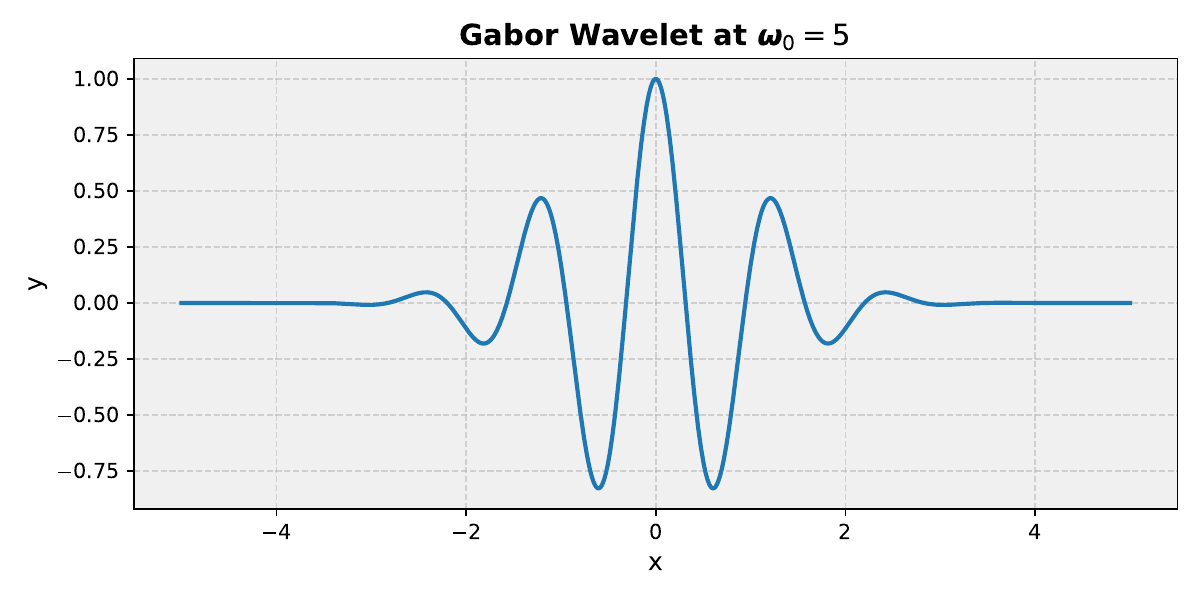}
    \caption{Gabor wavelet at $\omega_{0} =5$}
    \label{fig:gabor_3}
\end{subfigure}

\caption{Comparison between the Gabor wavelets at $\omega_{0} =3$ and $\omega_{0} =5$.}
\label{fig:gaborr}
\end{figure}

\textbf{Wavelet:} A function $\psi \in L^{2}(\mathbb{R})$ is referred to as a \emph{wavelet}, if it satisfies the admissibility criterion \cite{debnath2015wavelet}
\begin{equation}
C_{\psi} = \int_{-\infty}^{\infty} \frac{|\widehat{\psi}(\omega)|^{2}}{|\omega|} \, d\omega < \infty,
\label{eq:admissibility}
\end{equation}
where $\widehat{\psi}(\omega)$ represents the Fourier transform of $\psi(t)$. The condition $C_{\psi} < \infty$ ensures the invulnerability of the continuous wavelet transform. 
\cite{Uddin2023} employed the Mexican hat, Morlet, and Gaussian wavelets as activation functions, whereas softplus and \texttt{Tanh} remain among the most widely used traditional activation functions. Wavelet-based hyperbolic activation functions provide an alternative to the traditional \texttt{Tanh}-based activation functions. \\
\textbf{Mexican hat function:} The Mexican hat function can be defined as: 
\begin{equation}
\psi_{\text{MH}}(x)
= (1 - x^{2})\, \exp\!\left(-\ x^{2}/2\right).
\end{equation}
It is a real-valued, oscillatory, and localized activation function with good
space–frequency localization properties. The function satisfies
\[
\psi_{\text{MH}}(x) \to 0 \quad \text{as } |x| \to \infty,
\]
making it suitable for capturing local features in PINNs. Figure~\ref{fig:mexican_hat} shows the Mexican hat wavelet.\\
\textbf{Morlet functions:} The Morlet function can be defined as:
\begin{equation}
\psi_{\text{Morlet}}(x)
= \cos(\omega_{0} x)\, \exp\!\left(-\ x^{2}/2\right),
\end{equation}
where $\omega_{0}$ is a central frequency parameter. Figures~\ref{fig:morlet} provide the plot of the Morlet function.\\
\textbf{Hermite functions:} The proposed Hermite functions are constructed by drawing inspiration from the Hermite wavelet~\cite{debnath2015wavelet}. We can write the general form of the Hermite functions as  
\[
\psi_n(t) = H_n(t)\,\exp\!\left(-\ t^{2}/2\right), \qquad n = 0,1,2,\dots
\]
Figure~\ref{fig:hermite_wavelets} represents the plots of the first four
Hermite wavelets corresponding to \(n=1,2,3,\) and \(4\).
We can express the different orders of Hermite wavelets as follows:\\
\textbf{First order Hermite function:}
Using the first-order Hermite polynomial $H_{1}(x)=2x$, the Hermite function can be expressed as
\begin{equation}
\psi_{\text{Hermite-1}}(x)
= 2x\, \exp\!\left(-\ x^{2}/2\right).
\end{equation}
\textbf{Second-order Hermite function:} The Hermite function can be constructed from the second-order
Hermite polynomial $H_{2}(x) = 4x^{2} - 2$, giving
\begin{equation}
\psi_{\text{Hermite-2}}(x)
= (4x^{2} - 2)\, \exp\!\left(-\ x^{2}/2\right).
\end{equation}
\textbf{Third order Hermite functions:} Based on the third-order Hermite polynomial $H_{3}(x) = 8x^{3} - 12x$, the function can be expressed as
\begin{equation}
\psi_{\text{Hermite-3}}(x)
= (8x^{3} - 12x)\, \exp\!\left(-\ x^{2}/2\right).
\end{equation}
\textbf{Fourth order Hermite functions:} Using the fourth-order Hermite polynomial 
$H_{4}(x)=16x^{4}-48x^{2}+12$, the activation function is defined as
\begin{equation}
\psi_{\text{Hermite-4}}(x)
= \left(16x^{4} - 48x^{2} + 12\right) \exp\!\left(-\ x^{2}/2\right).
\end{equation}
\textbf{Gaussian functions:}
The Gaussian wavelet activation can be expressed as
\begin{equation}
\psi_{\text{Gauss}}(x)
= -x\exp\!\left(-\ x^{2}/2\right).
\end{equation}
In a similar manner, the Gaussian function inspired by the Gaussian wavelet is given by:
\begin{equation}
\psi_{\text{Gauss}}(x)
= \exp\!\left(-\ x^{2}/2\right).
\end{equation}
The plot in Figure~\ref{fig:gaussian} illustrates the Gaussian wavelet. In contrast to Gaussian wavelets, this work employs the Gaussian function. The plot in Figure~\ref{fig:gaussian_funtion} depicts the Gaussian function.\\
\textbf{Gabor functions:}
The complex Gabor wavelet is defined as
\[ 
\psi_{\text{complex}}(x)
= \exp\!\left(-\ x^{2}/(2\sigma^{2})\right)\,\exp\!\left(i\omega_{0} x\right),
\]
which can be written as
\[
\psi_{\text{complex}}(x)
= \exp\!\left(-\,x^{2}/(2\sigma^{2})\right)
\bigl[\cos(\omega_{0} x) + i \sin(\omega_{0} x)\bigr].
\]
Here, the real part is an even function and the imaginary part is an odd function.
The wavelet, therefore, forms an analytic, complex-valued representation.
The real Gabor wavelet is defined as
\[
\psi_{\text{real}}(x) 
= \exp\!\left(-\,x^{2}/(2\sigma^{2})\right)\,\cos(\omega_{0} x),
\]
where the Gaussian term \(\exp\!\left(-\,x^{2}/(2\sigma^{2})\right)\) provides frequency localization
and the cosine term introduces oscillations. This wavelet is purely real-valued
and exhibits even symmetry. The real Gabor wavelet is illustrated in Figure~\ref{fig:gabor_5} and \ref{fig:gabor_3}.

\begin{figure}[htbp]
\centering
\begin{subfigure}[b]{0.45\textwidth}
    \includegraphics[width=\textwidth]{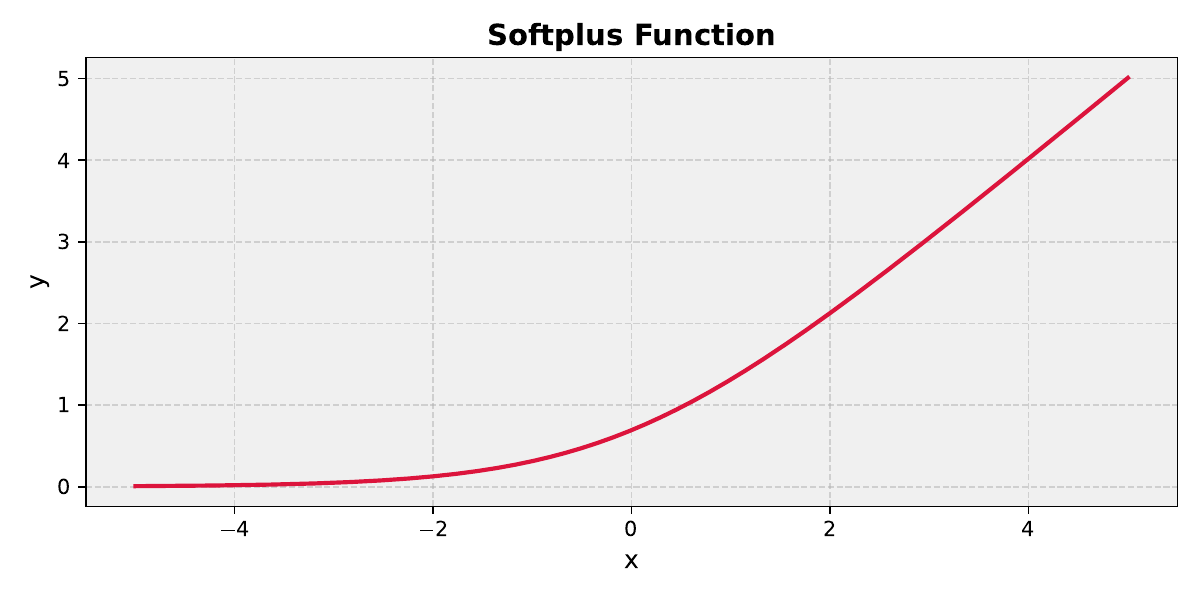}
    \caption{Softplus activation}
    \label{fig:softplus}
\end{subfigure}
\hfill
\begin{subfigure}[b]{0.45\textwidth}
    \includegraphics[width=\textwidth]{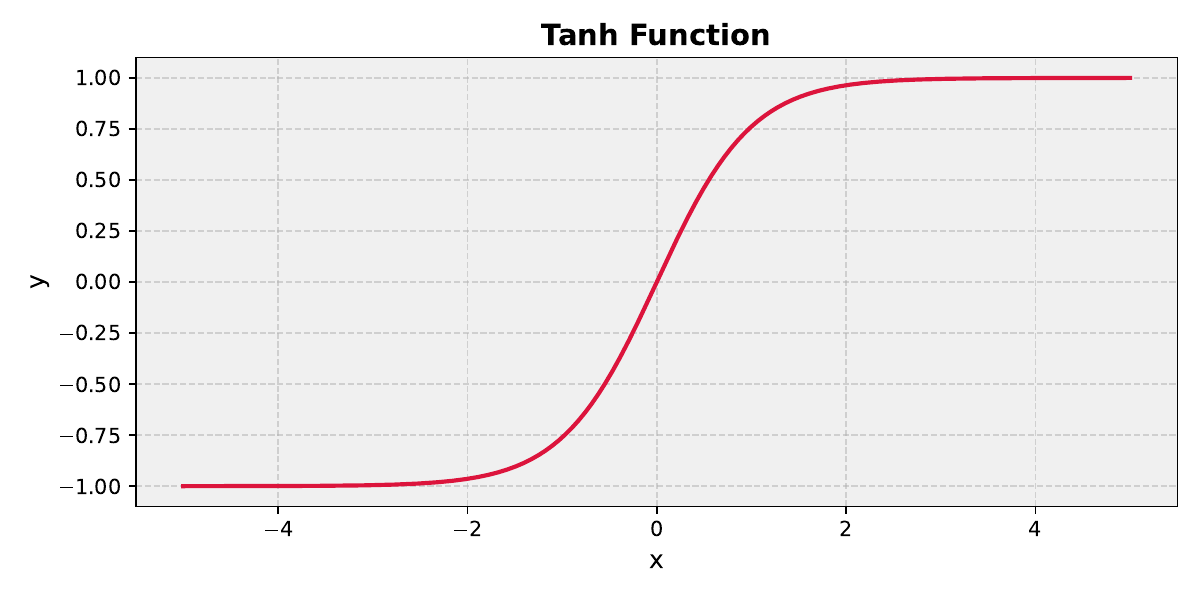}
    \caption{Hyperbolic tangent activation}
    \label{fig:tanh}
\end{subfigure}

\caption{Softplus and \texttt{Tanh} activation function.}
\label{fig:activations}
\end{figure}

\noindent
\textbf{Softplus function:}
The softplus activation function~\cite{dugas2000incorporating} is expressed as
\begin{equation}
\operatorname{softplus}(x)
= \ln\!\left( 1 +\exp\!\left(\,x\right) \right).
\end{equation}
The softplus function derivative yields the logistic sigmoid:
\begin{equation}
\frac{d}{dx}\operatorname{softplus}(x)
= \frac{1}{1 + \exp\!\left(\,-x\right)}.
\end{equation}
It is widely used due to its smoothness and stable gradient behavior. Figure~\ref{fig:softplus} illustrates the softplus activation function.\\
\textbf{Hyperbolic tangent function:}
The hyperbolic tangent represents a widely adopted nonlinear activation in NNs and is expressed as~\cite{dubey2022activation}
\begin{equation}
\operatorname{tanh}(x)
= \frac{\exp\!\left(\,x\right) - \exp\!\left(\,-x\right)}{\exp\!\left(\,x\right) +\exp\!\left(\,-x\right)}.
\end{equation}
The function is smooth, bounded, and symmetric, with range
\[
\operatorname{\tanh}(x) \in (-1,\,1).
\]
Its derivative is expressed as
\begin{equation}
\frac{d}{dx}\operatorname{\tanh}(x)
= 1 - \operatorname{\tanh}^{2}(x).
\end{equation}
The \(\operatorname{\tanh}(x)\) activation is centered at zero, which often
helps improve optimization and gradient flow compared to the sigmoid-type
activations. To simplify notation, the function $\tanh(x)$ or $\tanh(\beta x)$ is abbreviated as \texttt{Tanh}, as appropriate. Figure~\ref{fig:tanh} provides a depiction of the \texttt{Tanh} activation function.

\subsubsection{Newly designed custom activation functions}
The proposed activation functions are custom designs that blend wavelet-inspired components with a hyperbolic tangent nonlinearity, while incorporating softplus-parameterized trainable coefficients for improved adaptability. A distinctive feature of these wavelet-based activations is that their parameters are incorporated into the model’s set of learnable variables. As a result, the optimizer updates these coefficients during training, making them fully trainable. The activations are thus adaptive, since their shape evolves dynamically as the parameters are learned, and explicitly optimized through backpropagation and gradient-based learning.
\begin{definition}[\texttt{SoftMexTanh}]
Let $\alpha_{0}, \beta_{0}, \gamma_{0} \in \mathbb{R}$ be trainable parameters.  
Define their positive forms using the softplus function
\[
\alpha = \operatorname{softplus}(\alpha_{0}), \qquad
\beta  = \operatorname{softplus}(\beta_{0}), \qquad
\gamma = \operatorname{softplus}(\gamma_{0}),
\]
where
\[
\operatorname{softplus}(z) = \ln\!\left(1 + \exp\!\left(z\right)\right).
\]
Then the \emph{\texttt{SoftMexTanh} activation function} is formulated as
\[
\psi_{\text{\texttt{SoftMexTanh}}}(x)
=
\tanh(\beta x)\,
\left(1 - \gamma x^{2}\right)\,
\exp\!\left(-\alpha x^{2}\right).
\]
\end{definition}
\begin{definition}[\texttt{SoftMorTanh}]
Let $\omega_{0}, \sigma_{0} \in \mathbb{R}$ be trainable parameters.
Define their positive forms using the softplus function
\[
\omega = \operatorname{softplus}(\omega_{0}), \qquad
\sigma = \operatorname{softplus}(\sigma_{0}),
\]
where
\[
\operatorname{softplus}(z) = \ln\!\left(1 + \exp\!\left(z\right)\right).
\]
Then the \emph{\texttt{SoftMorTanh} activation function} is defined as
\[
\psi_{\text{\texttt{SoftMorTanh}}}(x)
=
\cos(\omega x)\,
\exp\!\left(-\frac{x^{2}}{2\sigma^{2}}\right)\,
\tanh(\beta x).
\]
\end{definition}
\begin{definition}[\texttt{SoftGaussTanh}]
Let $\alpha_{0}, \beta_{0} \in \mathbb{R}$ be trainable parameters.
Define their positive forms using the softplus function
\[
\alpha = \operatorname{softplus}(\alpha_{0}), \qquad
\beta  = \operatorname{softplus}(\beta_{0}),
\]
where
\[
\operatorname{softplus}(z) = \ln\!\left(1 + \exp\!\left(z\right)\right).
\]
Then the \emph{\texttt{SoftGaussTanh} activation function} is defined as
\[
\psi_{\text{\texttt{SoftGaussTanh}}}(x)
=
\,\tanh(\beta x)\;
\exp\!\left(-\alpha x^{2}\right).
\]
\end{definition}
\begin{definition}[\texttt{SoftGaborTanh}]
Let $\sigma_{0}, \omega_{0}, \beta_{0} \in \mathbb{R}$ be trainable parameters.
Define their positive forms using the softplus function
\[
\sigma = \operatorname{softplus}(\sigma_{0}), \qquad
\omega = \operatorname{softplus}(\omega_{0}), \qquad
\beta  = \operatorname{softplus}(\beta_{0}),
\]
where
\[
\operatorname{softplus}(z) = \ln\!\left(1 + \exp\!\left(z\right)\right).
\]
Let the Gaussian envelope and cosine carrier be defined by
\[
G(x) = \exp\!\left(-\frac{x^{2}}{2\sigma^{2}}\right), 
\qquad
C(x) = \cos(\omega x).
\]
Then the \emph{\texttt{SoftGaborTanh} activation function} activation is formulated as
\[
\psi_{\text{\texttt{SoftGaborTanh}}}(x)
=
\tanh(\beta x)\ \exp\!\left(-\frac{x^{2}}{2\sigma^{2}}\right)
\cos(\omega x).
\]
\end{definition}

\begin{definition}[\texttt{SoftHerTanh}]
Let $\alpha_{0}, \beta_{0} \in \mathbb{R}$ be trainable parameters and let $n$ denote the order of the Hermite polynomial.
Define the positive coefficients using the softplus function:
\[
\alpha = \operatorname{softplus}(\alpha_{0}), \qquad
\beta  = \operatorname{softplus}(\beta_{0}),
\]
where
\[
\operatorname{softplus}(z) = \ln\!\left(1 + \exp\!\left(z\right)\right).
\]
Let $H_{n}(x)$ denote the $n$th Hermite polynomial.  
As an illustration,
\[
H_{1}(x) = 2x, \qquad
H_{2}(x) = 4x^{2} - 2, \qquad
H_{3}(x) = 8x^{3} - 12x.
\]
Then the \emph{\texttt{SoftHerTanh} activation function} is formulated as
\[
\psi_{\text{\texttt{SoftHerTanh}}}(x)
=
\tanh(\beta x)\,
H_{n}(x)\,
\exp\!\left(-\alpha x^{2}\right).
\]
\end{definition}
\begin{remark}\label{remark:first}
In the proposed wavelet-based activation functions, parameters such as $\alpha$, $\beta$, and others are introduced and incorporated into the model’s set of learnable variables. Consequently, these coefficients are updated by the optimizer during training, making them fully trainable parameters. These activation functions become \textit{adaptive} because the shape of the activation evolves dynamically throughout training as the parameters are learned. All activation parameters are adaptively updated during training through standard gradient-based optimization. When the parameter of the \texttt{Tanh} term is fixed (i.e., $\beta$ is not trainable), the scaling of the $\tanh$ component remains unchanged throughout optimization. The corresponding activation functions, characterized by a trainable $\alpha$ and a fixed $\beta$, are denoted by the suffix \texttt{W}. Thus, the activations \texttt{SoftMexTanh}, \texttt{SoftMorTanh}, \texttt{SoftGaussTanh}, \texttt{SoftGaborTanh}, and \texttt{SoftHerTanh} without the trainable $\beta$
are denoted as \texttt{SoftMexTanhW}, \texttt{SoftMorTanhW}, \texttt{SoftGaussTanhW}, \texttt{SoftGaborTanhW} and \texttt{SoftHermTanhW} respectively.
\end{remark}
\begin{remark}
In certain cases, such as the convection problem, the \texttt{SoftMexTanh}, \texttt{SoftMorTanh}, \texttt{SoftGassTanh}, \texttt{SoftGaborTanh}, and \texttt{SoftHermTanh} activation functions do not yield sufficiently robust results. In such scenarios, we instead employ the corresponding variants with a trainable $\alpha$ and without a trainable $\beta$, namely \texttt{SoftMexTanhW}, \texttt{SoftMorTanhW}, \texttt{SoftGassTanhW}, \texttt{SoftGaborTanhW}, and \texttt{SoftHermTanhW}, respectively.
\end{remark}
\begin{remark}
For Hermite functions of orders one through four, the corresponding activation variants of 
\texttt{SoftHerTanh} and \texttt{SoftHerTanhW} are denoted as 
\texttt{SoftHer1Tanh} / \texttt{SoftHer1TanhW}, 
\texttt{SoftHer2Tanh} / \texttt{SoftHer2TanhW},
\texttt{SoftHer3Tanh} / \texttt{SoftHer3TanhW} and 
\texttt{SoftHer4Tanh} / \texttt{SoftHer4Tanh} respectively.
\end{remark}
\begin{remark}
For the \texttt{SoftGaborTanh} activation, the initial coefficient $\omega_{0}$ is set to the empirically optimal values of $3$ or $5$. For all other activation functions, the initial trainable parameters are set to 1.
\end{remark}

\section{Numerical Experiments} \label{sec:6}
All models are implemented in PyTorch~\cite{paszke2019pytorch} and are trained separately on a single \textbf{NVIDIA A100 GPU}. The experiments were conducted with a memory allocation of 32 GB. In numerical experiments, the first part describes the experiment setup, including dataset preparation, network architecture, and hyperparameter choices. The second part focuses on performance and error comparison. In this stage, results from different activation functions are evaluated using quantitative metrics, such as loss and error, and qualitative assessments via visual plots. This combined analysis clearly demonstrates the effect of activation choice on convergence behavior and prediction accuracy. The last section provides a visual summary of the error behavior using bar-plot analysis.
\subsection{Experiment Setup} 
This study evaluates performance across four representative PDEs classes, as outlined in Section~\ref{sec:3}. In the 1D cases, $N_{\textit{I}} = N_{\textit{B}} = 101$ uniformly spaced initial and boundary points are used, while the residual domain is represented by a $101 \times 101$ uniform grid, resulting in $N_{\textit{R}} = 10{,}201$ collocation points. The use of fewer training samples serves two objectives: enhancing computational efficiency and demonstrating the generalization performance of PINNs with limited data. Model evaluation is conducted on a $101\times 101$ uniform mesh defined over the residual domain. In the 2D Navier–Stokes equations, 2,500 collocation points are randomly sampled from the three-dimensional residual domain to train the model. The predictive performance is assessed by evaluating the estimated pressure field at the final time instance, $t = 20.0$. In the absence of an analytical reference solution for the 2D Navier–Stokes equations, the publicly released benchmark dataset documented in~\cite{raissi2019physics, Zhao2024PINNsFormer} is utilized. The data associated with the final time step are excluded from training and used instead as unseen test samples to examine the model’s generalization ability. Configuration settings for all test problems are chosen in accordance with \cite{raissi2017physics, raissi2019physics, krishnapriyan2021characterizing, Zhao2024PINNsFormer, wang2022and, gao2025ml}.

The training procedure employs the \texttt{L-BFGS} method with a strong Wolfe line search and is executed over 1000 iterations. For simplicity, the weighting coefficients in the optimization objective (Equation~\ref{eq:loss1}) are set as 
$\lambda_{\textit{R}} = \lambda_{\textit{I}} = \lambda_{\textit{B}} = 1$. In addition, the PINN training loss is reported for all experimental cases to assess convergence behavior. Experimental initialization employed a constant \textcolor{orange}{random seed value of 5} to support reproducibility. For quantitative evaluation, we employ standard error metrics widely used in prior studies~\cite{raissi2019physics, krishnapriyan2021characterizing, mcclenny2020self, Zhao2024PINNsFormer}, namely the relative Mean Absolute Error (\texttt{rMAE}) and the relative Root Mean Square Error (\texttt{rRMSE}), corresponding to the relative $\ell_{1}$ and $\ell_{2}$ errors, respectively. The error can be defined as follows:
\begin{equation}
\begin{gathered}
    \texttt{rMAE}   =  \frac{\sum\limits_{n=1}^N |\hat{u}(x_n,t_n)-u(x_n,t_n)|}{\sum\limits_{n=1}^{N_{\textit{res}}}|u(x_n,t_n)|}\\
    \texttt{rRMSE}  = \sqrt{\frac{\sum\limits_{n=1}^N |\hat{u}(x_n,t_n)-u(x_n,t_n)|^2}{\sum\limits_{n=1}^N|u(x_n,t_n)|^2}},
\end{gathered}    
\end{equation}
here, $N$ denotes the total number of test points, $\hat{u}$ represents the model-predicted solution and $u$ corresponds to the reference solution. In addition, we include the training PINNs loss for all cases. The \textcolor{red}{highest} and \textcolor{yellow}{lowest} error values are highlighted for clarity. Table~\ref{tbl:hyper} presents the principal hyperparameters and optimization settings adopted in this study.
\begin{table}[h!]
\centering
\renewcommand{\arraystretch}{1.25}
\begin{tabular}{lll}
\hline\hline
\textbf{Category} & \textbf{Configuration} & \textbf{Value} \\ 
\hline
\multirow{3}{*}{Model Hyper-parameters} 
& Hidden layers       & 4 \\
& Hidden size         & 512 \\
& Loss weights        & $\lambda_{R} = \lambda_{I} = \lambda_{B} = 1$ \\
\hline
\multirow{2}{*}{Training Settings} 
& Optimizer           & L-BFGS (Strong Wolfe) \\
& Iterations          & 1000 \\
\hline\hline
\end{tabular}
\caption{Model hyperparameter and training configurations used for the PINNs model.}
\label{tbl:hyper}
\end{table}
\subsection{Performance and Error Comparison}
\begin{figure}[h!]
    \centering
    \includegraphics[width=0.45\textwidth]{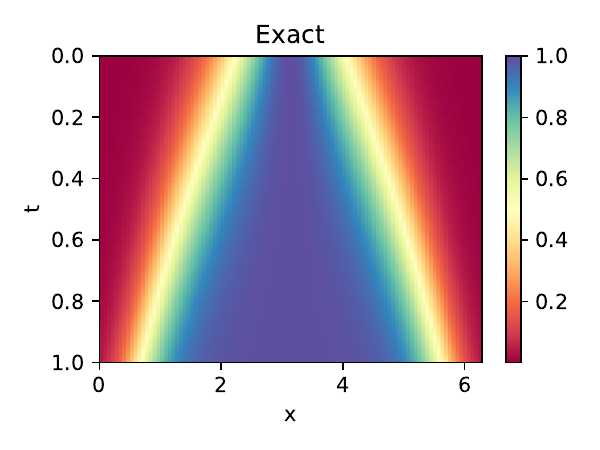}
    \caption{Exact solution}
    \label{fig:reaction_exact}
\end{figure}

\begin{figure}[htbp]
\centering
\begin{subfigure}[b]{0.32\textwidth}
    \includegraphics[width=\textwidth]{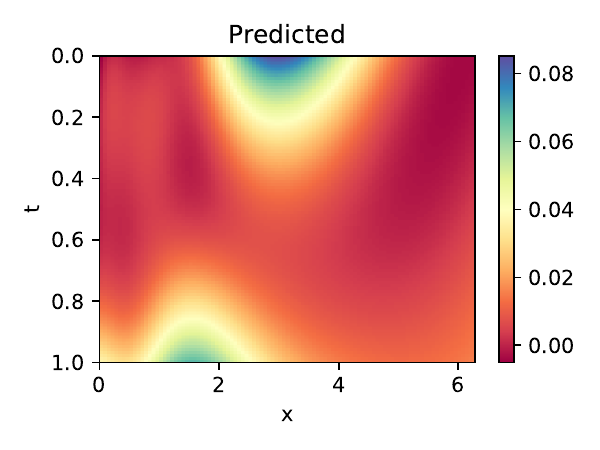}
    \caption{PINNs prediction:\texttt{Tanh}}
    \label{fig:reaction_pred}
\end{subfigure}
\hfill
\begin{subfigure}[b]{0.32\textwidth}
    \includegraphics[width=\textwidth]{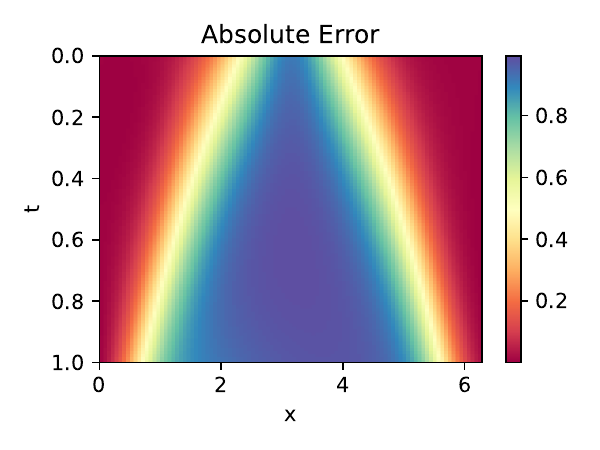}
    \caption{Absolute error:\texttt{Tanh}}
    \label{fig:reaction_error}
\end{subfigure}
\caption{Predicted solution and absolute error using \texttt{Tanh} for the 1D reaction equation.}
\label{fig:reaction_comparison}
\end{figure}

\begin{figure}[htbp]
\centering
\begin{subfigure}[b]{0.32\textwidth}
    \includegraphics[width=\textwidth]{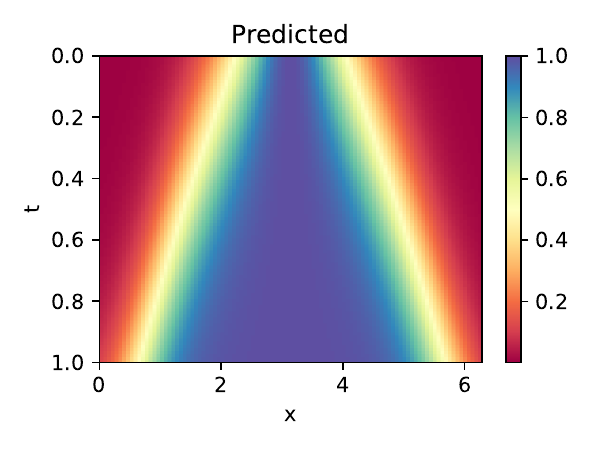}
    \caption{PINNs prediction:\texttt{SoftMaxTanh}}
    \label{fig:1dreaction_pinns_pred_max}
\end{subfigure}
\hfill
\begin{subfigure}[b]{0.32\textwidth}
    \includegraphics[width=\textwidth]{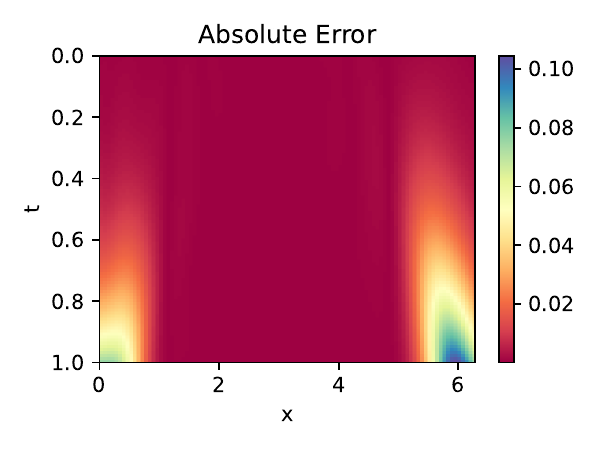}
    \caption{Absolute error:\texttt{SoftMaxTanh}}
    \label{fig:1dreaction_pinns_max_err}
\end{subfigure}
\caption{Predicted solution and absolute error using \texttt{SoftMaxTanh} for the 1D reaction equation.}
\label{fig:reaction_comparison_max}
\end{figure}

\begin{figure}[htbp]
\centering
\begin{subfigure}[b]{0.32\textwidth}
    \includegraphics[width=\textwidth]{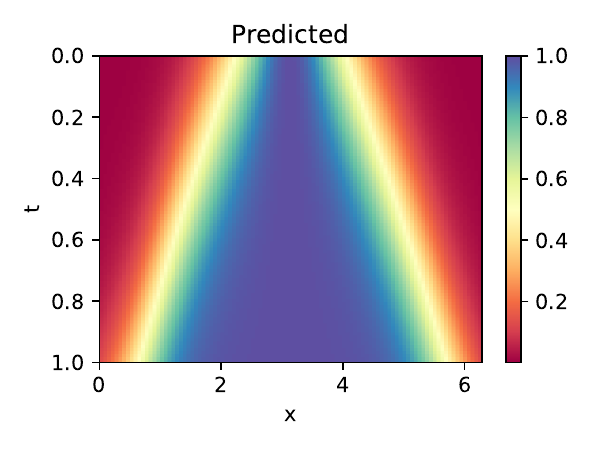}
    \caption{PINNs prediction:\texttt{SoftMorTanh}}
    \label{fig:1dreaction_pinns_pred_mor}
\end{subfigure}
\hfill
\begin{subfigure}[b]{0.32\textwidth}
    \includegraphics[width=\textwidth]{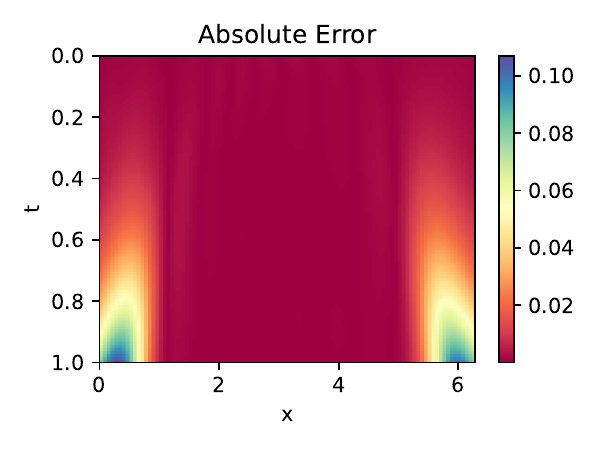}
    \caption{Absolute error:\texttt{SoftMorTanh}}
    \label{fig:1dreaction_pinns_mor_err}
\end{subfigure}
\caption{Predicted solution and absolute error using \texttt{SoftMorTanh} for the 1D reaction equation.}
\label{fig:reaction_comparison_mor}
\end{figure}

\begin{figure}[htbp]
\centering
\begin{subfigure}[b]{0.32\textwidth}
    \includegraphics[width=\textwidth]{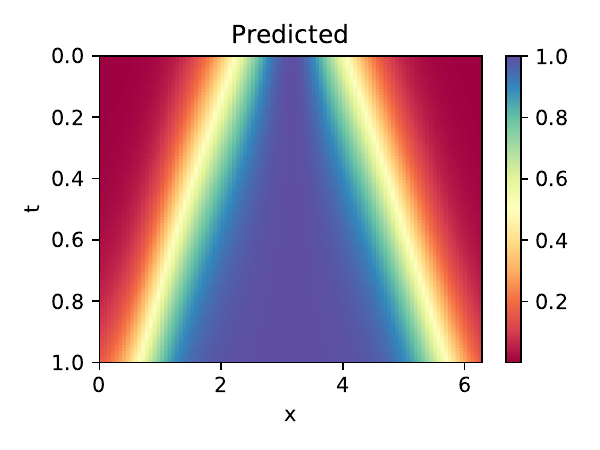}
    \caption{PINNs prediction:\texttt{SoftGaussTanh}}
    \label{fig:1dreaction_pinns_pred_gauss}
\end{subfigure}
\hfill
\begin{subfigure}[b]{0.32\textwidth}
    \includegraphics[width=\textwidth]{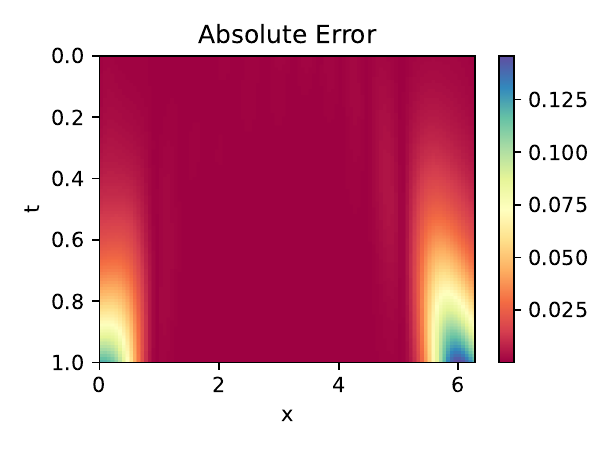}
    \caption{Absolute error:\texttt{SoftGaussTanh}}
    \label{fig:1dreaction_pinns_gauss_err}
\end{subfigure}
\caption{Predicted solution and absolute error using \texttt{SoftGaussTanh} for the 1D reaction equation.}
\label{fig:reaction_comparison_gauss}
\end{figure}

\begin{figure}[htbp]
\centering
\begin{subfigure}[b]{0.32\textwidth}
    \includegraphics[width=\textwidth]{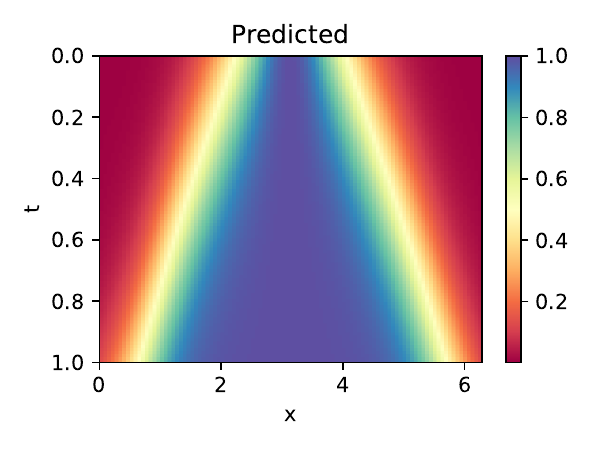}
    \caption{PINNs prediction:\texttt{SoftHerTanh}}
    \label{fig:1dreaction_pinns_pred_her}
\end{subfigure}
\hfill
\begin{subfigure}[b]{0.32\textwidth}
    \includegraphics[width=\textwidth]{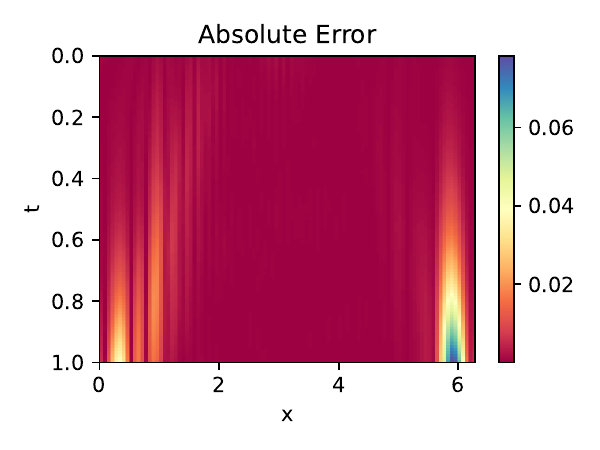}
    \caption{Absolute error:\texttt{SoftHerTanh}}
    \label{fig:1dreaction_pinns_her_err}
\end{subfigure}
\caption{Predicted solution and absolute error using \texttt{SoftHerTanh} for the 1D reaction equation.}
\label{fig:reaction_comparison_her}
\end{figure}

\begin{figure}[htbp]
\centering
\begin{subfigure}[b]{0.32\textwidth}
    \includegraphics[width=\textwidth]{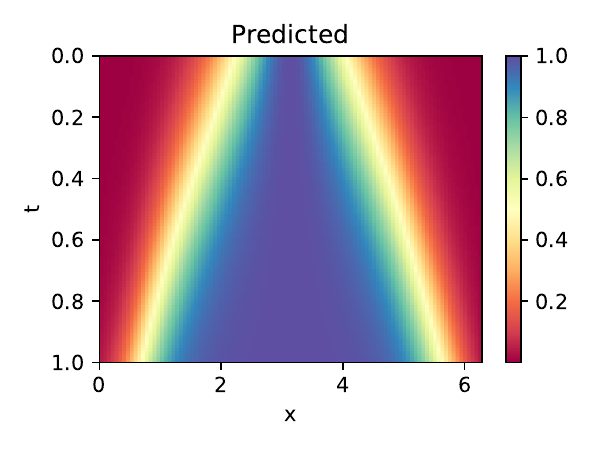}
    \caption{PINN prediction:\texttt{SoftGaborTanh}}
    \label{fig:1dreaction_pinns_pred_gab}
\end{subfigure}
\hfill
\begin{subfigure}[b]{0.32\textwidth}
    \includegraphics[width=\textwidth]{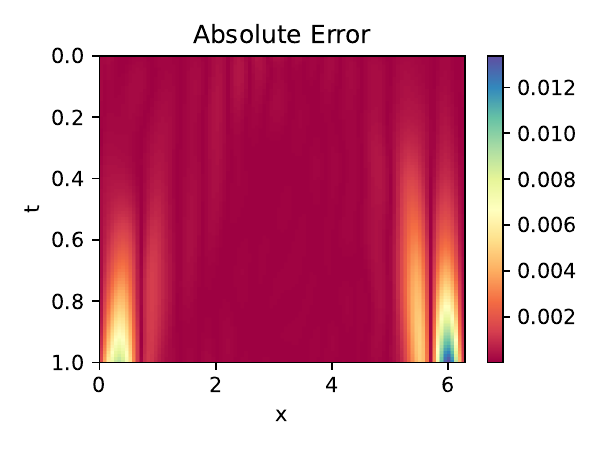}
    \caption{Absolute error:\texttt{SoftGaborTanh}}
    \label{fig:1dreaction_pinns_her_gab}
\end{subfigure}
\caption{Predicted solution and absolute error using \texttt{SoftGaborTanh} for the 1D reaction equation.}
\label{fig:reaction_comparison_gab}
\end{figure}

We compared the performance, loss, and error distributions using contour plots and quantitative tables. The four PDE simulations are as follows:\\
\textbf{1D Reaction Equation:} In Section~\ref{sec:3}, the formulation given in equation~\ref{eqn:1dreactio} is employed. Figure~\ref{fig:reaction_exact} presents the ground-truth solution, while Figures~\ref{fig:reaction_pred} and~\ref{fig:reaction_error} show the corresponding PINNs prediction and absolute error obtained using the standard \texttt{Tanh} activation function~\cite{raissi2019physics}. Similarly, Figures~\ref{fig:1dreaction_pinns_pred_max} and~\ref{fig:1dreaction_pinns_max_err} depict the predicted solution and absolute error achieved using the proposed activation function~\texttt{SoftMaxTanh}. In the same manner, Figures~\ref{fig:1dreaction_pinns_pred_mor} and~\ref{fig:1dreaction_pinns_mor_err} illustrate the performance of activation function~\texttt{SoftMorTanh}, while Figures~\ref{fig:1dreaction_pinns_pred_gauss} and~\ref{fig:1dreaction_pinns_gauss_err} show the results for activation function~\texttt{SoftGaussTanh}. Likewise, Figures~\ref{fig:1dreaction_pinns_pred_her} and~\ref{fig:1dreaction_pinns_her_err} present the outcomes for activation function~\texttt{SoftHer3Tanh} and Figures~\ref{fig:1dreaction_pinns_pred_gab} and~\ref{fig:1dreaction_pinns_her_gab} display the corresponding prediction and error for activation function~\texttt{SoftGaborTanh}. 
The visual comparisons clearly demonstrate that the proposed activation functions yield more accurate predictions and lower errors than conventional \texttt{Tanh}-based PINNs. Furthermore, Table~\ref{table_4} provides a detailed comparison of training loss and relative error. Table~\ref{table_4} presents the loss and error results for the 1D reaction equation. From Table~\ref{table_44} we observe that, relative to the standard \texttt{Tanh} activation, the proposed activation functions \texttt{SoftGaussTanh}, \texttt{SoftMorTanh}, \texttt{SoftMexTanh}, \texttt{SoftHer3Tanh}, and \texttt{SoftGaborTanh} achieve progressively lower error values in descending order, demonstrating superior approximation capability.


\begin{figure}[h!]
    \centering
    \includegraphics[width=0.45\textwidth]{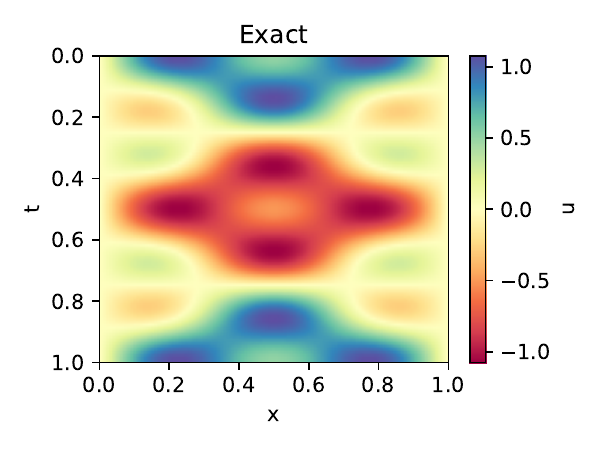}
    \caption{Exact solution}
    \label{fig:wave_exact}
\end{figure}

\begin{figure}[htbp]
\centering
\begin{subfigure}[b]{0.32\textwidth}
    \includegraphics[width=\textwidth]{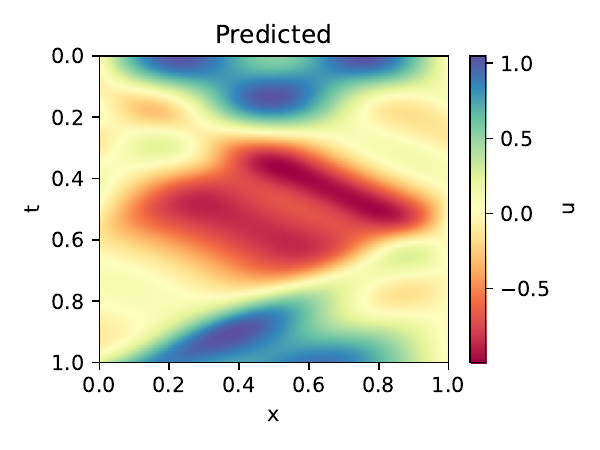}
    \caption{PINNs:\texttt{Tanh}}
    \label{fig:wave_pred}
\end{subfigure}
\hfill
\begin{subfigure}[b]{0.32\textwidth}
    \includegraphics[width=\textwidth]{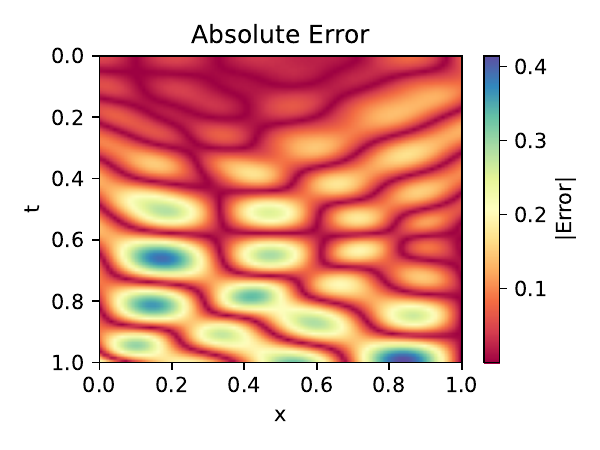}
    \caption{Absolute error:\texttt{Tanh}}
    \label{fig:wave_error}
\end{subfigure}

\caption{Predicted solution and absolute error using \texttt{Tanh} for the 1D wave equation.
}
\label{fig:wave_comparison}
\end{figure}

\begin{figure}[htbp]
\centering
\begin{subfigure}[b]{0.32\textwidth}
    \includegraphics[width=\textwidth]{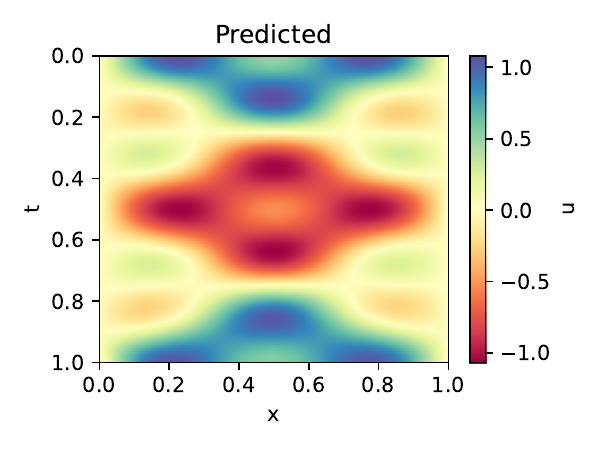}
    \caption{PINNs:\texttt{SoftMaxTanh}}
    \label{fig:wave_pred_max}
\end{subfigure}
\hfill
\begin{subfigure}[b]{0.32\textwidth}
    \includegraphics[width=\textwidth]{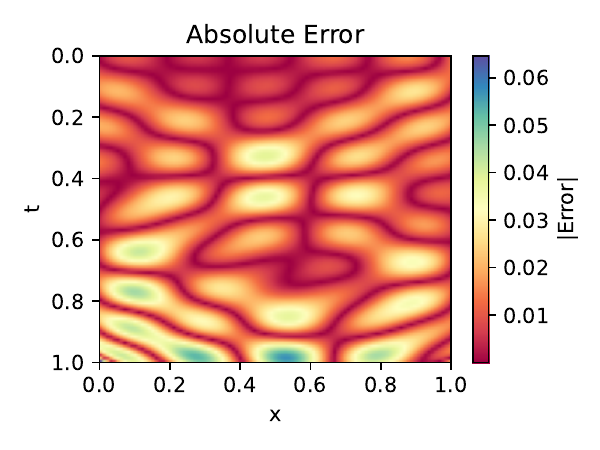}
    \caption{Absolute error:\texttt{SoftMaxTanh}}
    \label{fig:wave_error_max}
\end{subfigure}
\caption{Predicted solution and absolute error using \texttt{SoftMaxTanh} for the 1D wave equation.}
\label{fig:wave_comparison_max}
\end{figure}

\begin{figure}[htbp]
\centering
\begin{subfigure}[b]{0.32\textwidth}
    \includegraphics[width=\textwidth]{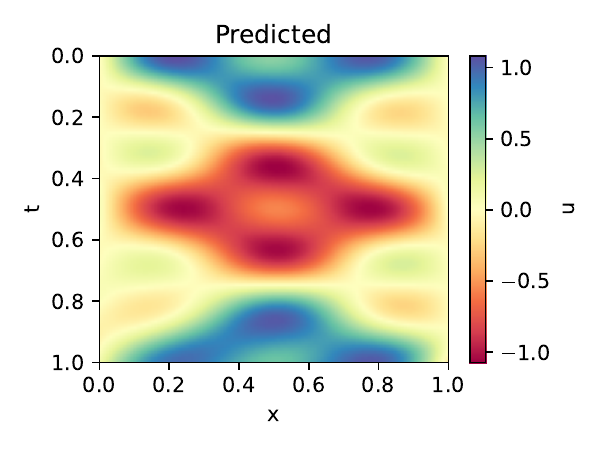}
    \caption{PINNs:\texttt{SoftMorTanh}}
    \label{fig:wave_pred_mor}
\end{subfigure}
\hfill
\begin{subfigure}[b]{0.32\textwidth}
    \includegraphics[width=\textwidth]{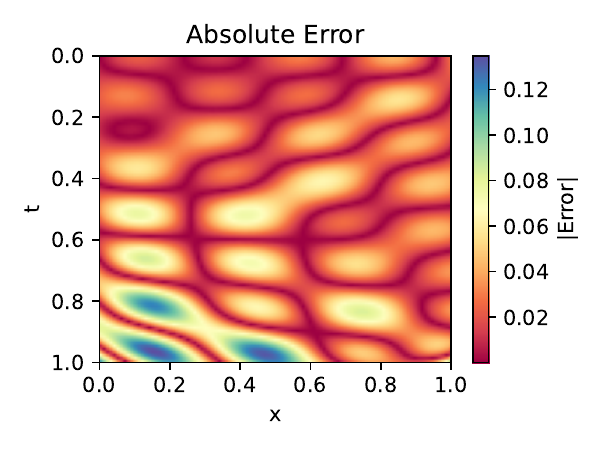}
    \caption{Absolute error:\texttt{SoftMorTanh}}
    \label{fig:wave_error_mor}
\end{subfigure}
\caption{Predicted solution and absolute error using \texttt{SoftMorTanh} for the 1D wave equation.}
\label{fig:wave_comparison_mor}
\end{figure}

\begin{figure}[htbp]
\centering
\begin{subfigure}[b]{0.32\textwidth}
    \includegraphics[width=\textwidth]{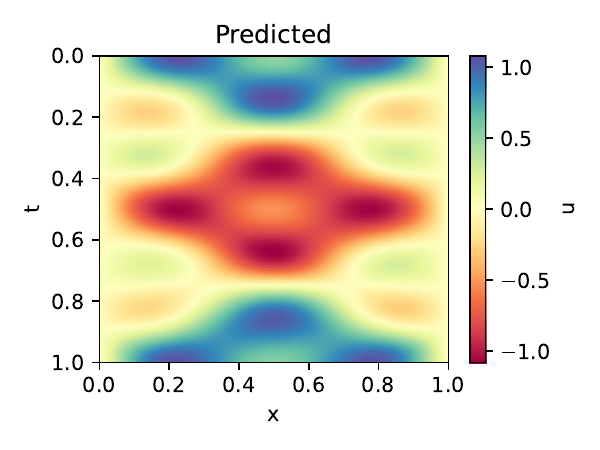}
    \caption{PINNs:\texttt{SoftGaussTanh}}
    \label{fig:wave_pred_gauss}
\end{subfigure}
\hfill
\begin{subfigure}[b]{0.32\textwidth}
    \includegraphics[width=\textwidth]{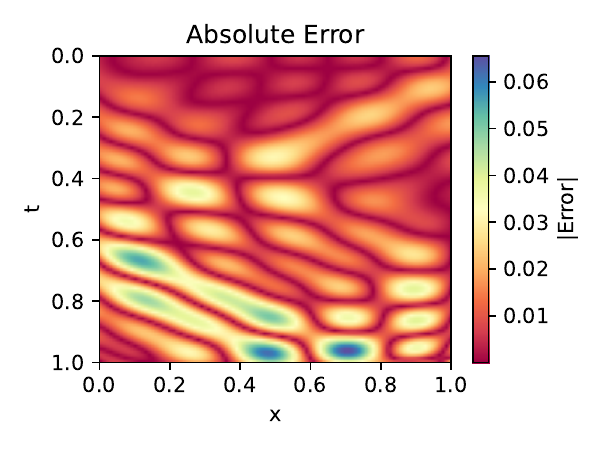}
    \caption{Absolute error:\texttt{SoftGaussTanh}}
    \label{fig:wave_error_gauss}
\end{subfigure}
\caption{Predicted solution and absolute error using \texttt{SoftGaussTanh} for the 1D wave equation.}
\label{fig:wave_comparison_gauss}
\end{figure}

\begin{figure}[htbp]
\centering
\begin{subfigure}[b]{0.32\textwidth}
    \includegraphics[width=\textwidth]{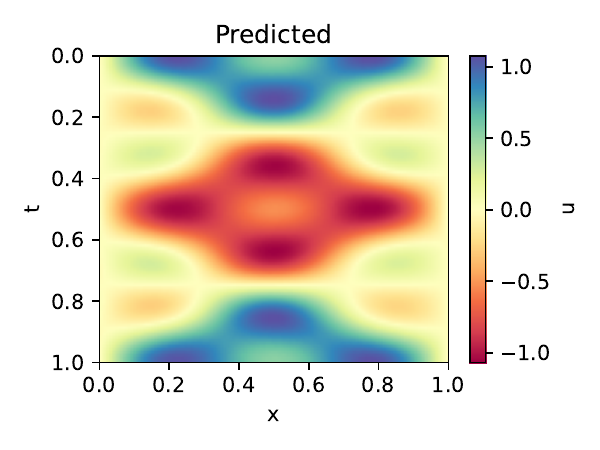}
    \caption{PINNs:\texttt{SoftHer2Tanh}}
    \label{fig:wave_pred_herm}
\end{subfigure}
\hfill
\begin{subfigure}[b]{0.32\textwidth}
    \includegraphics[width=\textwidth]{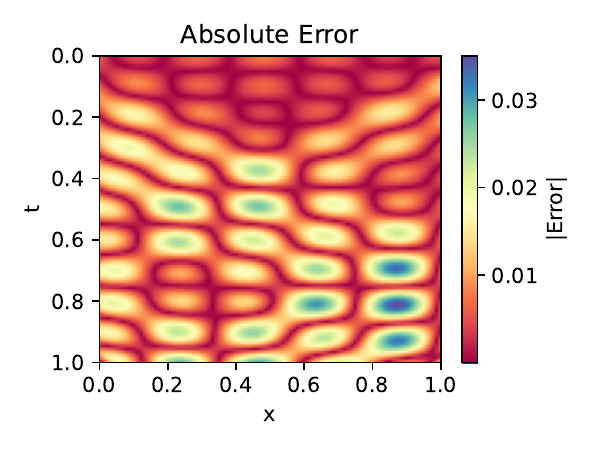}
    \caption{Absolute error:\texttt{SoftHer2Tanh}}
    \label{fig:wave_error_herm}
\end{subfigure}
\caption{Predicted solution and absolute error using \texttt{SoftHer2Tanh} for the 1D wave equation.}
\label{fig:wave_comparison_herm}
\end{figure}

\begin{figure}[htbp]
\centering
\begin{subfigure}[b]{0.32\textwidth}
    \includegraphics[width=\textwidth]{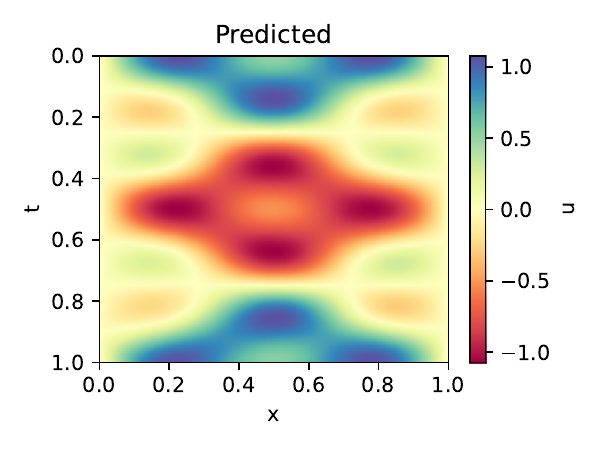}
    \caption{PINNs:\texttt{SoftGaborTanh}}
    \label{fig:wave_pred_gab}
\end{subfigure}
\hfill
\begin{subfigure}[b]{0.32\textwidth}
    \includegraphics[width=\textwidth]{figures/1dwave_gauss_error.pdf}
    \caption{Absolute error:\texttt{SoftGaborTanh}}
    \label{fig:wave_error_gab}
\end{subfigure}
\caption{Predicted solution and absolute error using \texttt{SoftGaborTanh} for the 1D wave equation.}
\label{fig:wave_comparison_gab}
\end{figure}

\begin{table}[!h]
\centering
\resizebox{0.95\textwidth}{!}{%
\begin{tabular}{||c|c|c c c c c||} 
\hline
\diagbox[width=5.5cm]{\texttt{Loss / Error}}{\textbf{Activation functions}} & 
\textbf{Tanh\cite{raissi2019physics}} & \textbf{SoftMexTanh} & \textbf{SoftMorTanh} & \textbf{SoftGaussTanh} & \textbf{SoftHer3Tanh} & \textbf{SoftGaborTanh} \\ [0.5ex] 
\hline
\texttt{Loss} & \cellcolor{red}\textbf{1.99e-01} & 2.90e-06 & 6.02e-06  & 3.92e-06 & 9.16e-06  & \cellcolor{yellow}\textbf{7.00e-07 } \\ 
\texttt{rMAE} & \cellcolor{red}\textbf{0.975} & 0.011 & 0.0153 & 0.015 & 0.004 & \cellcolor{yellow}\textbf{0.0010} \\ 
\texttt{rRMSE} & \cellcolor{red}\textbf{0.973} & 0.023 & 0.0295 & 0.032 & 0.010 & \cellcolor{yellow}\textbf{0.0021} \\ 
\hline
\end{tabular}%
}
\caption{Loss and error comparison across different activation functions for the 1D reaction equation.}
\label{table_4}
\end{table}
\noindent
\textbf{1D Wave Equation:} In Section~\ref{sec:3}, the formulation given in equation~\ref{eqn:1dwave} is employed. Figure~\ref{fig:wave_exact} depicts the reference or ground-truth solution, whereas Figures~\ref{fig:wave_pred} and~\ref{fig:wave_error} illustrate the corresponding PINN-based prediction and its absolute error using the conventional \texttt{Tanh} activation function~\cite{raissi2019physics}. In contrast, Figures~\ref{fig:wave_pred_max} and~\ref{fig:wave_error_max} present the predicted solution and associated error when employing the proposed activation function~ \texttt{SoftMaxTanh}. Likewise, Figures~\ref{fig:wave_pred_mor} and~\ref{fig:wave_error_mor} demonstrate the behavior of activation function~\texttt{SoftMorTanh}, while Figures~\ref{fig:wave_pred_gauss} and~\ref{fig:wave_error_gauss} report the outcomes obtained with activation function~\texttt{SoftGaussTanh}. Similarly, Figures~\ref{fig:wave_pred_herm} and~\ref{fig:wave_error_herm} display the results for activation function~\texttt{SoftHerTanh} and Figures~\ref{fig:wave_pred_gab} and~\ref{fig:wave_error_gab} show the corresponding predictions and absolute error for activation function~\texttt{SoftGaborTanh}. From the comparative visual analysis, it is evident that the proposed activation functions achieve higher prediction accuracy and significantly reduced errors compared to the standard \texttt{Tanh}-based PINN model. Moreover, Table~\ref{table_44} summarizes the training loss and relative error for all tested activation functions. From Table~\ref{table_44}, it can be observed that, relative to the standard \texttt{Tanh} activation, the proposed activation functions \texttt{SoftMorTanh}, \texttt{SoftGaussTanh}, \texttt{SoftGaborTanh}, \texttt{SoftMexTanh}, and \texttt{SoftHermTanh} yield progressively lower error values in descending order, thereby demonstrating superior approximation capability for the 1D wave equation.

\begin{figure}[htbp]
    \centering
    \includegraphics[width=0.45\textwidth]{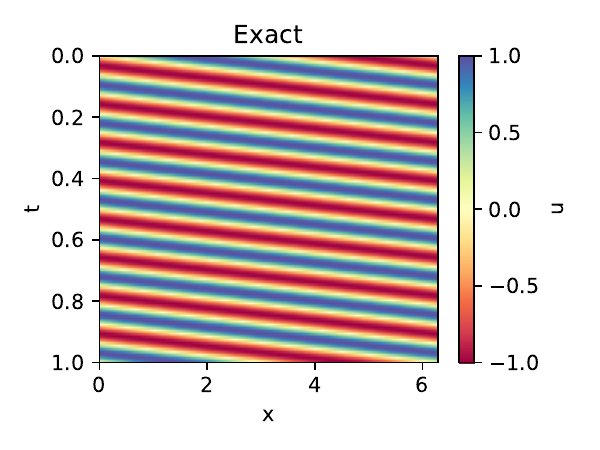}
    \caption{Exact solution}
    \label{fig:convection_exact}
\end{figure}

\begin{figure}[htbp]
\centering
\begin{subfigure}[b]{0.32\textwidth}
    \includegraphics[width=\textwidth]{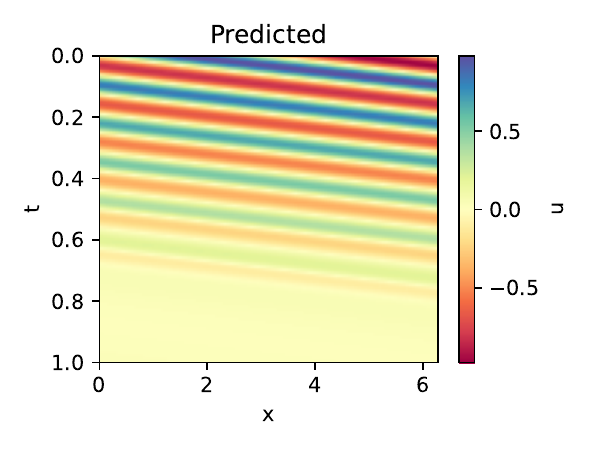}
    \caption{PINNs prediction:\texttt{Tanh}}
    \label{fig:convection_pred}
\end{subfigure}
\hfill
\begin{subfigure}[b]{0.32\textwidth}
    \includegraphics[width=\textwidth]{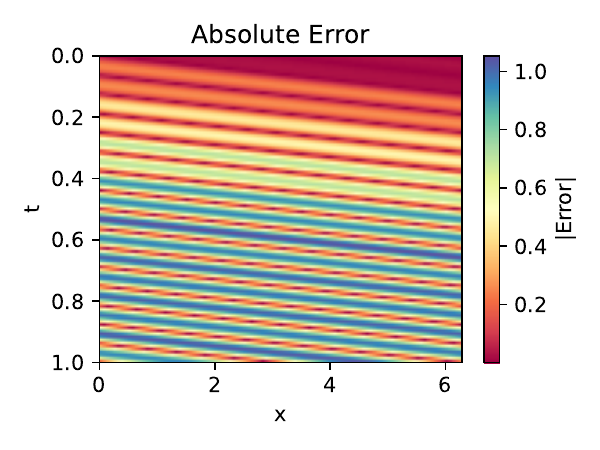}
    \caption{Absolute error:\texttt{Tanh}}
    \label{fig:convection_error}
\end{subfigure}
\caption{Predicted solution and absolute error using \texttt{Tanh} for the 1D convection.}
\label{fig:convection_comparison}
\end{figure}

\begin{figure}[htbp]
\centering
\begin{subfigure}[b]{0.32\textwidth}
    \includegraphics[width=\textwidth]{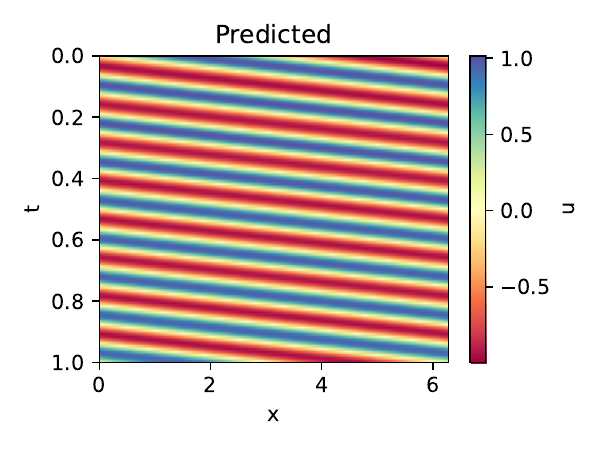}
    \caption{PINNs prediction:\texttt{SoftGaborTanhW}}
    \label{fig:convection_pred_gab}
\end{subfigure}
\hfill
\begin{subfigure}[b]{0.32\textwidth}
    \includegraphics[width=\textwidth]{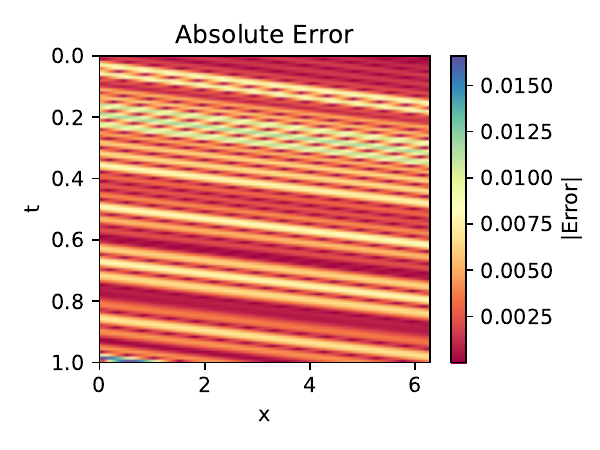}
    \caption{Absolute error:\texttt{SoftGaborTanhW}}
    \label{fig:convection_pred_error}
\end{subfigure}
\caption{Predicted solution and absolute error using \texttt{SoftGaborTanhW} for the 1D convection.}
\label{fig:convection_pred_comparision}
\end{figure}

\begin{table}[!h]
\centering
\resizebox{0.95\textwidth}{!}{%
\begin{tabular}{||c|c|c c c c c||} 
\hline
\diagbox[width=5.5cm]{\texttt{Loss / Error}}{\textbf{Activation functions}} & 
\textbf{Tanh\cite{raissi2019physics}} & \textbf{SoftMexTanh} & \textbf{SoftMorTanh} & \textbf{SoftGaussTanh} & \textbf{SoftGaborTanh} & \textbf{SoftHer2Tanh} \\ [0.5ex] 
\hline
\texttt{Loss} & \cellcolor{red}\textbf{1.0e-02} & 5.0e-04 & 2.5e-03  & 4.1e-04 & 2.1e-04  & \cellcolor{yellow}\textbf{2.2e-04}  \\ 
\texttt{rMAE} & \cellcolor{red}\textbf{0.214} & 0.030 & 0.066 & 0.015 & 0.029 & \cellcolor{yellow}\textbf{0.018} \\ 
\texttt{rRMSE} & \cellcolor{red}\textbf{0.223} & 0.030 & 0.069 & 0.032 & 0.031 & \cellcolor{yellow}\textbf{0.019} \\ 
\hline
\end{tabular}%
}
\caption{Loss and error comparison of PINNs with different activation functions for the 1D wave equation.}
\label{table_44}
\end{table}
\noindent
\textbf{Convection Equation:} In Section~\ref{sec:3}, the formulation given in equation~\ref{eqn:con} is employed. Figure~\ref{fig:convection_exact} illustrates the reference (ground-truth) solution, while Figures~\ref{fig:convection_pred} and~\ref{fig:convection_error} present the corresponding PINNs predictions and absolute errors obtained using the conventional \texttt{Tanh} activation function~\cite{raissi2019physics}. In contrast, Figures~\ref{fig:convection_pred_gab} and~\ref{fig:convection_pred_error} show the predicted field and error distribution generated by the proposed activation function~ \texttt{SoftGaborTanh}. Incremental improvements over \texttt{Tanh} are observed for the remaining proposed activation functions. Consequently, all proposed activations except \texttt{SoftGaborTanh} exhibit prediction accuracy and error levels comparable to \texttt{Tanh}. A comparative examination of the visual results clearly indicates that the proposed activation functions yield more accurate approximations with considerably lower absolute errors than the baseline \texttt{Tanh}-based PINNs. In addition, Table~\ref{table_444} provides a comprehensive summary of the training loss and relative error values, reinforcing the enhanced convergence rate and predictive performance achieved by the proposed activation functions.

\begin{figure}[htbp]
\centering
\includegraphics[width=0.45\textwidth]{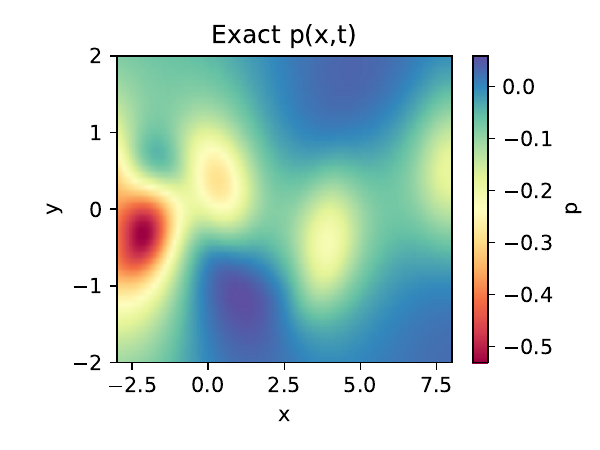}
\caption{Ground truth solution of 2D Navier-Stokes Equations}
\label{fig:navier_exact}
\end{figure}

\begin{figure}[htbp]
\centering
\begin{subfigure}[b]{0.32\textwidth}
    \includegraphics[width=\textwidth]{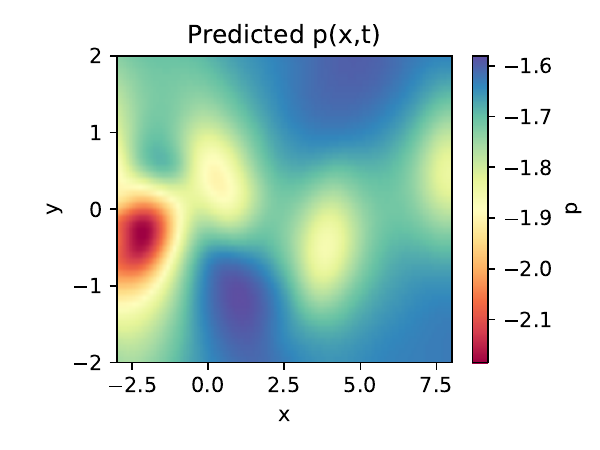}
    \caption{PINNs prediction:\texttt{Tanh}}
    \label{fig:navier_pred}
\end{subfigure}
\hfill
\begin{subfigure}[b]{0.32\textwidth}
    \includegraphics[width=\textwidth]{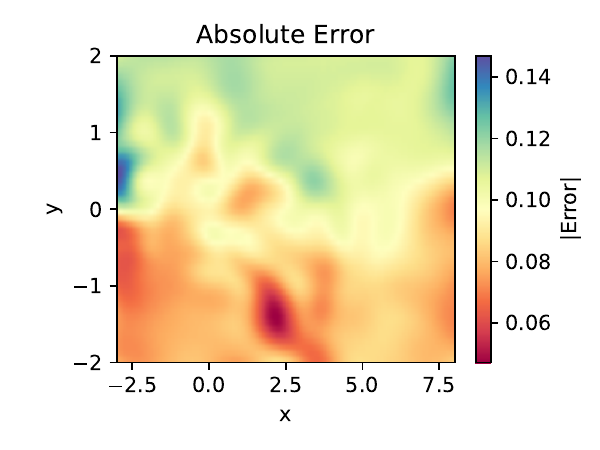}
    \caption{Absolute error:\texttt{Tanh}}
    \label{fig:navier_error}
\end{subfigure}

\caption{PINNs prediction and absolute error for the 2D Navier–Stokes equation using \texttt{Tanh} activation.}
\label{fig:navier_comparison}
\end{figure}

\begin{figure}[htbp]
\centering
\begin{subfigure}[b]{0.32\textwidth}
    \includegraphics[width=\textwidth]{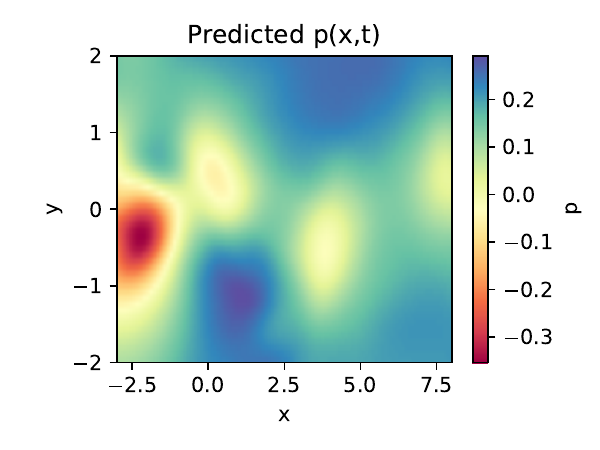}
    \caption{PINNs prediction:\texttt{SoftMaxTanh}}
    \label{fig:navier_pred_max}
\end{subfigure}
\hfill
\begin{subfigure}[b]{0.32\textwidth}
    \includegraphics[width=\textwidth]{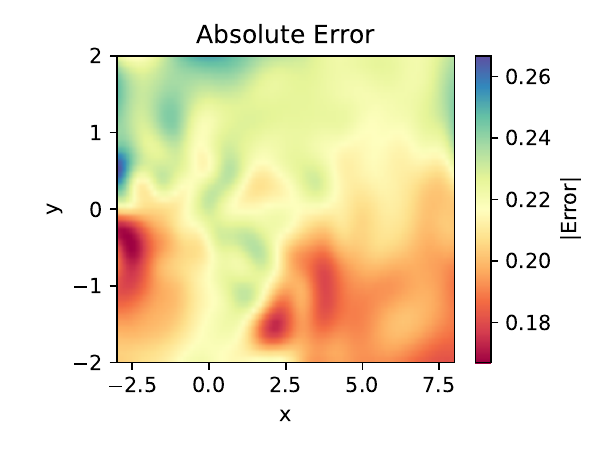}
    \caption{Absolute error: \texttt{SoftMaxTanh}}
    \label{fig:navier_error_max}
\end{subfigure}
\caption{PINNs prediction and absolute error for the 2D Navier–Stokes equation using  \texttt{SoftMaxTanh} activation.}
\label{fig:navier_comparison_max}
\end{figure}

\begin{figure}[htbp]
\centering
\begin{subfigure}[b]{0.32\textwidth}
    \includegraphics[width=\textwidth]{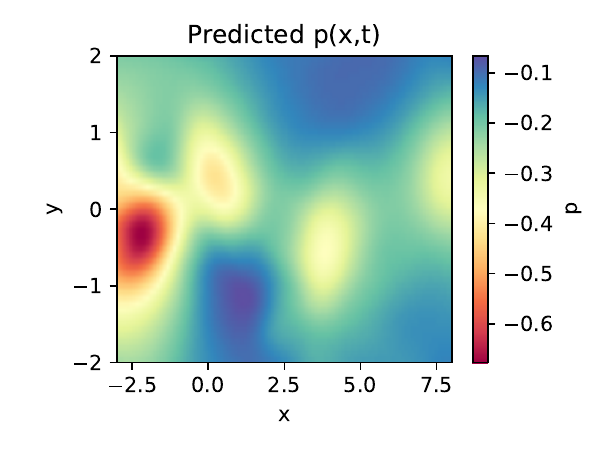}
    \caption{PINNs prediction:\texttt{SoftMorTanh}}
    \label{fig:navier_pred_mor}
\end{subfigure}
\hfill
\begin{subfigure}[b]{0.32\textwidth}
    \includegraphics[width=\textwidth]{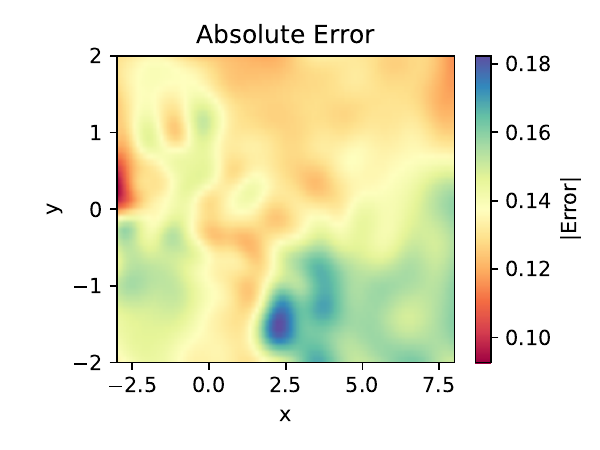}
    \caption{Absolute error:\texttt{SoftMorTanh}}
    \label{fig:navier_error_mor}
\end{subfigure}
\caption{PINNs prediction and absolute error for the 2D Navier–Stokes equation using  \texttt{SoftMorTanh} activation.}
\label{fig:navier_comparison_mor}
\end{figure}

\begin{figure}[htbp]
\centering
\begin{subfigure}[b]{0.32\textwidth}
    \includegraphics[width=\textwidth]{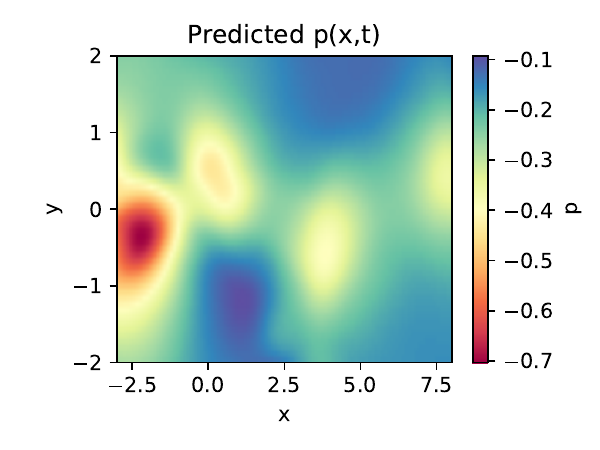}
    \caption{PINNs prediction:\texttt{SoftGaussTanh}  }
    \label{fig:navier_pred_gauss}
\end{subfigure}
\hfill
\begin{subfigure}[b]{0.32\textwidth}
    \includegraphics[width=\textwidth]{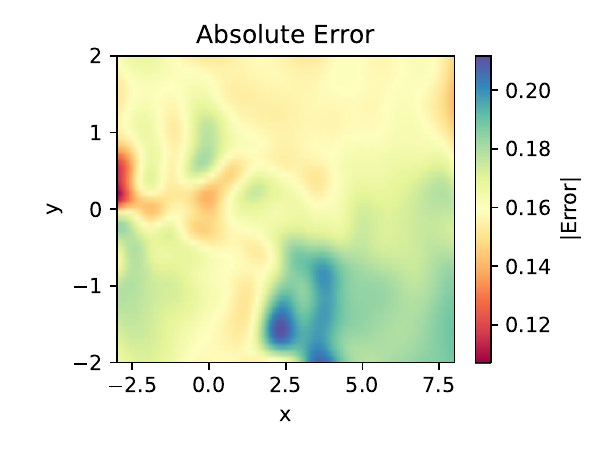}
    \caption{Absolute error:\texttt{SoftGaussTanh}}
    \label{fig:navier_error_gauss}
\end{subfigure}
\caption{PINNs prediction and absolute error for the 2D Navier–Stokes equation using  \texttt{SoftGaussTanh} activation.}
\label{fig:navier_comparison_gauss}
\end{figure}

\begin{figure}[htbp]
\centering
\begin{subfigure}[b]{0.32\textwidth}
    \includegraphics[width=\textwidth]{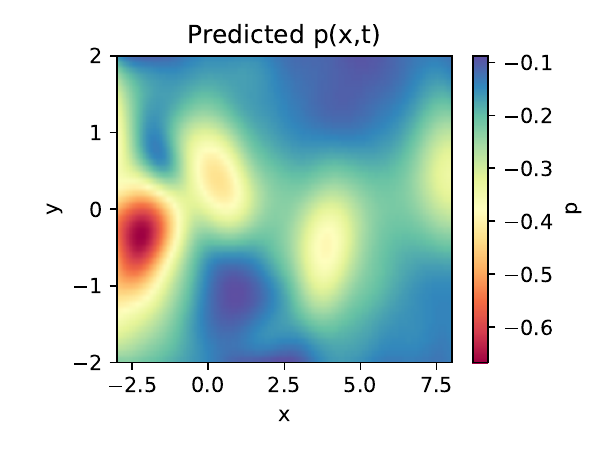}
    \caption{PINNs prediction:\texttt{SoftHer2Tanh}}
    \label{fig:navier_pred_her}
\end{subfigure}
\hfill
\begin{subfigure}[b]{0.32\textwidth}
    \includegraphics[width=\textwidth]{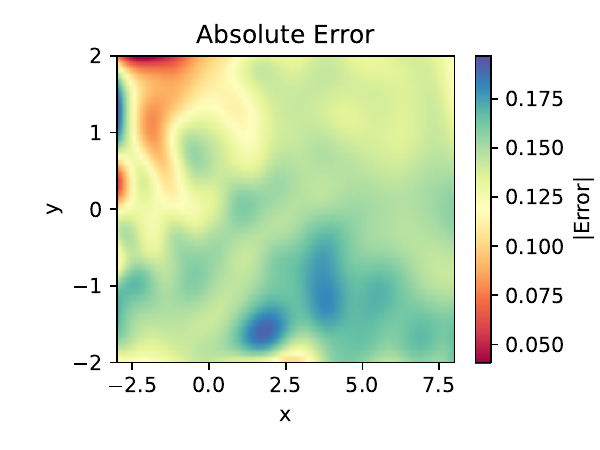}
    \caption{Absolute error:\texttt{SoftHer2Tanh}}
    \label{fig:navier_error_her}
\end{subfigure}
\caption{PINNs prediction and absolute error for the 2D Navier–Stokes equation using  \texttt{SoftHer2Tanh} activation.}
\label{fig:navier_comparison_her}
\end{figure}

\begin{figure}[htbp]
\centering
\begin{subfigure}[b]{0.32\textwidth}
    \includegraphics[width=\textwidth]{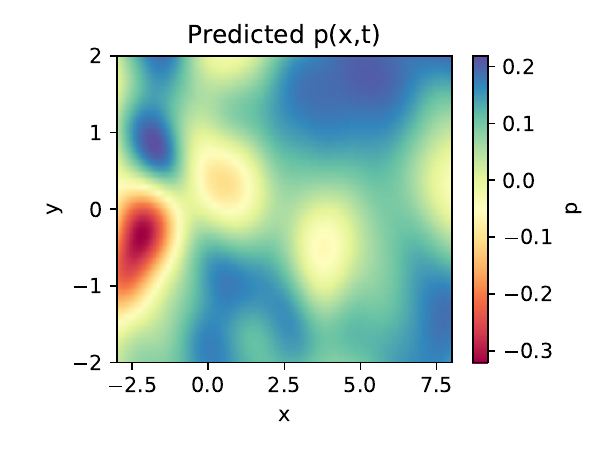}
    \caption{PINNs prediction:\texttt{SoftGaborTanh}}
    \label{fig:navier_pred_gabor}
\end{subfigure}
\hfill
\begin{subfigure}[b]{0.32\textwidth}
    \includegraphics[width=\textwidth]{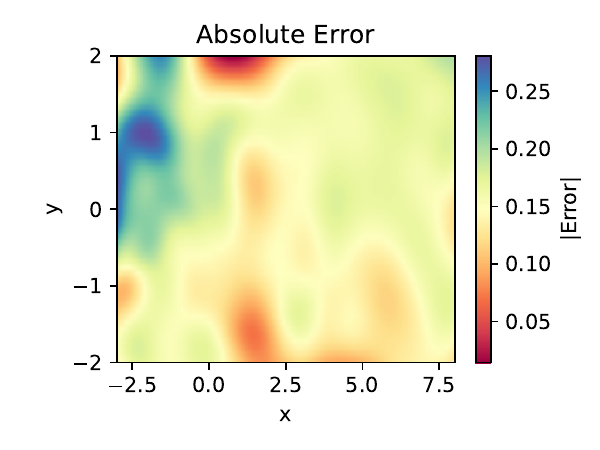}
    \caption{Absolute error:\texttt{SoftGaborTanh}}
    \label{fig:navier_error_gabor}
\end{subfigure}
\caption{PINNs prediction and absolute error for the 2D Navier–Stokes equation using  \texttt{SoftGaborTanh} activation.}
\label{fig:navier_comparison_gabor}
\end{figure}

\begin{table}[!h]
\centering
\resizebox{0.95\textwidth}{!}{%
\begin{tabular}{||c|c|c c c c c||} 
\hline
\diagbox[width=5.5cm]{\texttt{Loss / Error}}{\textbf{Activation functions}} & 
\textbf{Tanh\cite{raissi2019physics}} & \textbf{SoftMexTanhW} & \textbf{SoftMorTanhW} & \textbf{SoftGaussTanhW} & \textbf{SoftHer3TanhW} & \textbf{SoftGaborTanhW} \\ [0.5ex] 
\hline
\texttt{Loss} & \cellcolor{red}\textbf{0.0163} & 0.0069 & 0.00699 & 0.0106 & 0.010 & \cellcolor{yellow}\textbf{4.167e-05} \\ 
\texttt{rMAE} & \cellcolor{red}\textbf{0.724} & 0.4619 & 0.4763 & 0.6543 & 0.6473 & \cellcolor{yellow}\textbf{0.00606} \\ 
\texttt{rRMSE} & \cellcolor{red}\textbf{0.796}& 0.5321 & 0.5475 & 0.7336 & 0.7260 & \cellcolor{yellow}\textbf{0.0068} \\ 
\hline
\end{tabular}%
}
\caption{Loss and error comparison of PINNs with different activation functions for the 1D convection equation.}
\label{table_444}
\end{table}
\noindent
\textbf{2D Navier–Stokes Equations:} The formulation presented in Section~\ref{sec:3} together with Equation~\ref{eqn:ns} is employed. The PINNs demonstrate a good qualitative match to the ground-truth (reference) solution of the equations, yet the quantitative relative error is considerably high. Figure~\ref{fig:navier_exact} shows the reference pressure field. Figures~\ref{fig:navier_pred} and~\ref{fig:navier_error} present the corresponding PINN predictions and absolute error distributions obtained with a \texttt{Tanh}~\cite{raissi2019physics}. In comparison, Figures~ \ref{fig:navier_pred_max} and~ \ref{fig:navier_error_max} present the predicted flow field and error contours produced by the proposed activation function \texttt{SoftMaxTanh}. Likewise, Figures~\ref{fig:navier_pred_mor} and~ \ref{fig:navier_error_mor} display the performance of activation function~\texttt{SoftMorTanh}, while Figures~\ref{fig:navier_pred_gauss} and~\ref{fig:navier_error_gauss} demonstrate the corresponding results for activation function~ \texttt{SoftGaussTanh}. Moreover, Figures~\ref{fig:navier_pred_her} and~\ref{fig:navier_error_her} show the predictions obtained with activation function~\texttt{SoftHer2Tanh} and Figures~\ref{fig:navier_pred_gabor} and~\ref{fig:navier_error_gabor} illustrate the results achieved with activation function~\texttt{SoftGaborTanh}.

The comparative visual analysis indicates that the introduced activation functions improve the quality of the predicted velocity and pressure fields and reduce errors relative to the standard \texttt{Tanh}-based PINN. Furthermore, Table~\ref{table_4444} summarizes the training loss and relative error metrics, confirming the superior convergence behavior and improved generalization performance of the proposed activation functions. Table~\ref{table_4444} shows that, relative to the standard \texttt{Tanh} activation, the proposed activation functions \texttt{SoftMexTanh}, \texttt{SoftGaussTanh}, \texttt{SoftGaborTanh}, \texttt{SoftHer2Tanh}, and \texttt{SoftMorTanh} yield progressively lower error values in descending order, thereby demonstrating superior approximation capability. An additional analysis is conducted by excluding trainable parameters from the \texttt{Tanh} term, following the discussion in Remark~\ref{remark:first}. Table~\ref{table_44444} illustrates that, relative to the standard \texttt{Tanh} activation, the proposed activation functions \texttt{SoftGaussTanhW}, \texttt{SoftMexTanhW}, \texttt{SoftHer2TanhW}, \texttt{SoftGaborTanhW}, and \texttt{SoftMorTanhW} achieve progressively lower error values, thereby demonstrating superior approximation capability.


\begin{table}[!h]
\centering
\resizebox{0.95\textwidth}{!}{%
\begin{tabular}{||c|c|c c c c c||} 
\hline
\diagbox[width=5.5cm]{\texttt{Loss / Error}}{\textbf{Activation functions}} & 
\textbf{Tanh \cite{raissi2019physics}} & \textbf{SoftMexTanh} & \textbf{SoftMorTanh} & \textbf{SoftGaussTanh} & \textbf{SoftGaborTanh} & \textbf{SoftHer2Tanh} \\ [0.5ex] 
\hline
\texttt{Loss} & \cellcolor{red}\textbf{1.101e-04} & 7.07e-06  & \cellcolor{yellow}\textbf{7.85e-06} & 1.09e-05 & 5.34e-06 & 1.04e-06 \\ 
\texttt{rMAE} & \cellcolor{red}\textbf{17.80} & 2.31 & \cellcolor{yellow}\textbf{1.515} & 1.807 & 1.713 & 1.547 \\ 
\texttt{rRMSE} & \cellcolor{red}\textbf{12.35}& 1.60 & \cellcolor{yellow}\textbf{1.05} & 1.258 & 1.217 & 1.084 \\ 
\hline
\end{tabular}%
}
\caption{Loss and error comparison of PINNs with different activation functions for the 2D Navier–Stokes equations.}
\label{table_4444}
\end{table}

\begin{table}[!h]
\centering
\resizebox{0.95\textwidth}{!}{%
\begin{tabular}{||c|c|c c c c c||} 
\hline
\diagbox[width=5.5cm]{\texttt{Loss / Error}}{\textbf{Activation functions}} & 
\textbf{Tanh \cite{raissi2019physics}} & \textbf{SoftMexTanhW} & \textbf{SoftMorTanhW} & \textbf{SoftGaussTanhW} & \textbf{SoftGaborTanhW} & \textbf{SoftHer2TanhW} \\ [0.5ex] 
\hline
\texttt{Loss} & \cellcolor{red}\textbf{1.101e-04} & 7.07e-06  & \cellcolor{yellow}\textbf{9.61e-06} & 8.601e-05 & 6.17e-06 & 7.145e-06 \\ 
\texttt{rMAE} & \cellcolor{red}\textbf{17.800} & 2.00 & \cellcolor{yellow}\textbf{0.539} & 4.097 & 0.835 & 1.52 \\ 
\texttt{rRMSE} & \cellcolor{red}\textbf{12.356}& 1.398 & \cellcolor{yellow}\textbf{0.388} & 2.845 & 0.590 & 1.06 \\ 
\hline
\end{tabular}%
}
\caption{Loss and error comparison of PINNs with different activation functions for the 2D Navier–Stokes equations.}
\label{table_44444}
\end{table}

\begin{remark}
We can observe from Tables~\ref{table_4444} and~\ref{table_44444} that removing the trainable parameter from the hyperbolic tangent component of the proposed activation functions leads to improved relative $\ell_{1}$ and $\ell_{2}$ errors, although the overall loss increases slightly.  
\end{remark}

\begin{remark}
We also compare our results with those reported in~\cite{Zhao2024PINNsFormer}. For the 1D reaction equation, 1D wave equation, convection equation, and 2D Navier–Stokes equations, the \texttt{rRMSE} values reported in~\cite{Zhao2024PINNsFormer}, obtained using the same hyperparameters as in our experiments, are 0.021, 0.488, 0.647, and 8.03 (3.0 at zero seed , the result agrees with~\cite{gao2025ml}), respectively. The training time for the 1D reaction equation demonstrates a clear efficiency advantage of the proposed approach: PINNs with the \texttt{Tanh} activation achieve 14.03\,it/s, PINNsFormer reaches only 6.54\,it/s, whereas the best-performing variant using the proposed activation function, PINN-\texttt{SoftGaborTanh}, attains 9.01\,it/s. Although PINNsFormer requires more memory than the standard PINN formulation, its training time remains higher than that of the proposed framework, and the resulting accuracy is inferior to the results obtained in this study. Similar to classical PINNs~\cite{raissi2019physics}, other methods such as FLS~\cite{wong2022learning} and QRes~\cite{bu2021quadratic} struggle to achieve high-accuracy approximations to PDE solutions when using standard activation functions. Our comparative study demonstrates that the proposed framework delivers significantly lower prediction errors than these existing approaches~\cite{raissi2019physics, wong2022learning, bu2021quadratic, Zhao2024PINNsFormer}. PINNs~\cite{raissi2019physics}, QRes~\cite{bu2021quadratic} and FLS~\cite{wong2022learning} yield almost identical results, while the methods in~\cite{Xu2025SubSequential} and~\cite{gao2025ml} produce similar errors. PINNsFormer \cite{Zhao2024PINNsFormer}, PINNMamba\cite{Xu2025SubSequential}, and ML-PINN \cite{gao2025ml} require significantly more memory and training time compared to standard PINNs (see \cite{Xu2025SubSequential} and \cite{gao2025ml}).
\end{remark}

\subsection{Discussions}
\begin{figure}[htbp]
\centering

\begin{subfigure}[b]{0.8\textwidth}
    \centering
    \includegraphics[width=\textwidth]{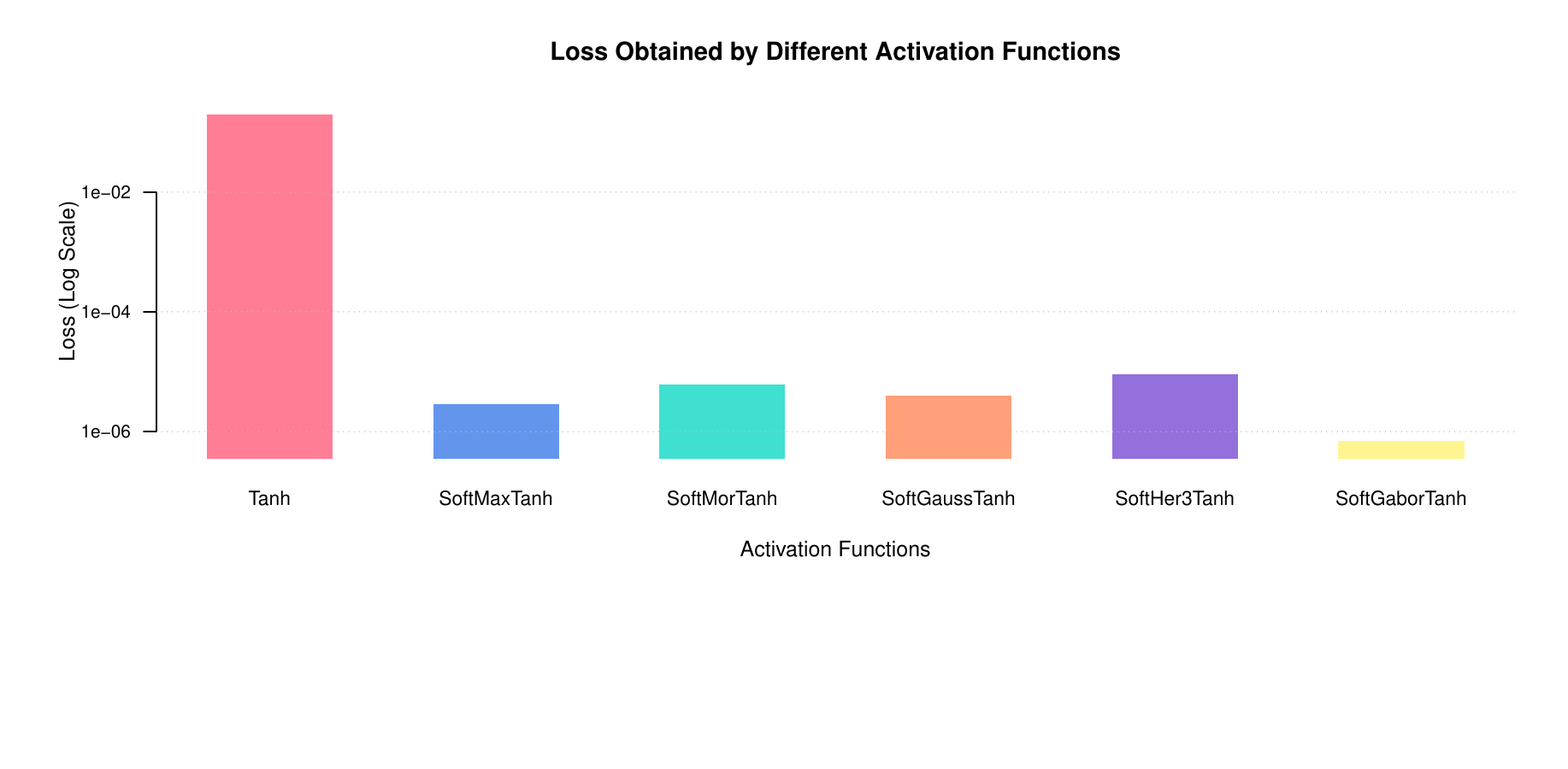}
    \caption{Loss values for different activation functions.}
    \label{fig:loss_plot_1d_reaction}
\end{subfigure}

\vspace{0.4cm}

\begin{subfigure}[b]{0.8\textwidth}
    \centering
    \includegraphics[width=\textwidth]{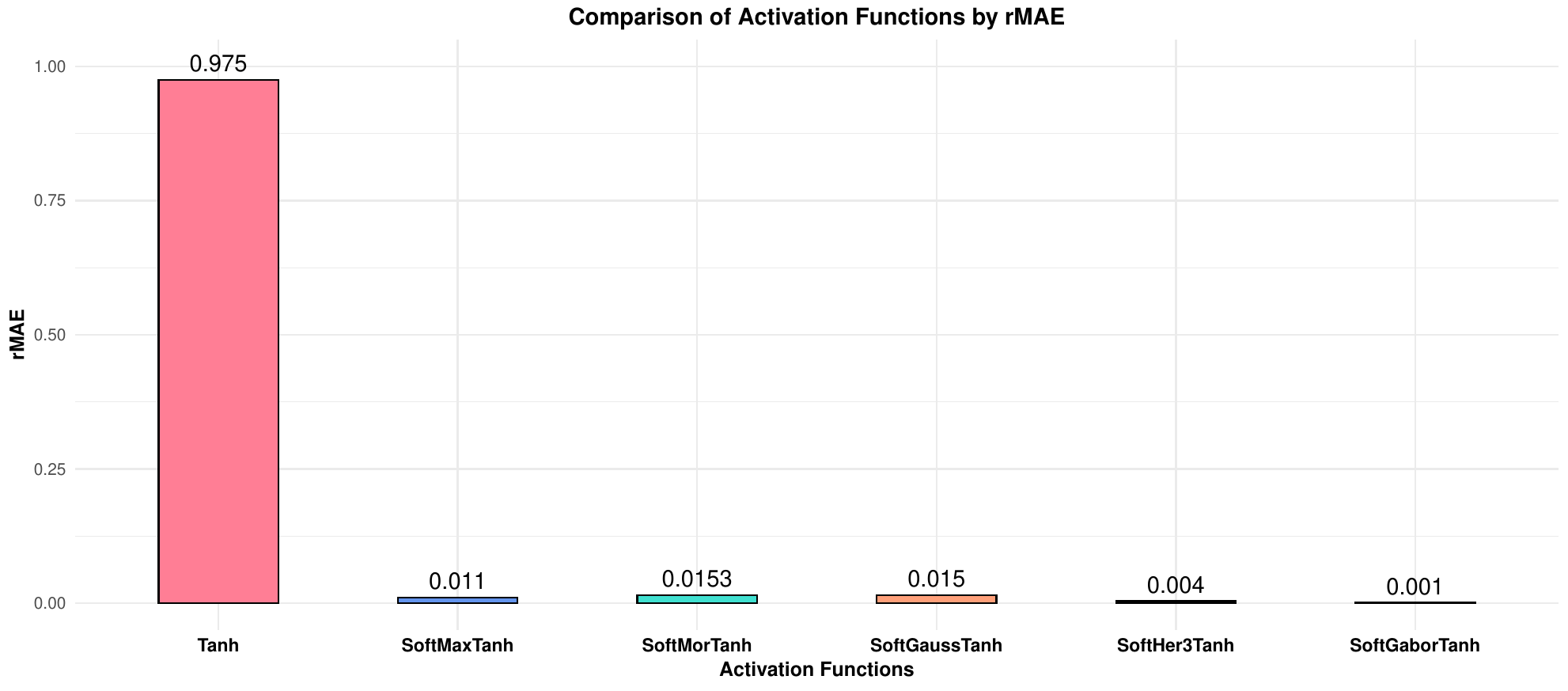}
    \caption{Relative MAE (rMAE) for different activation functions.}
    \label{fig:rMAE_act_1dreaction}
\end{subfigure}

\vspace{0.4cm}

\begin{subfigure}[b]{0.8\textwidth}
    \centering
    \includegraphics[width=\textwidth]{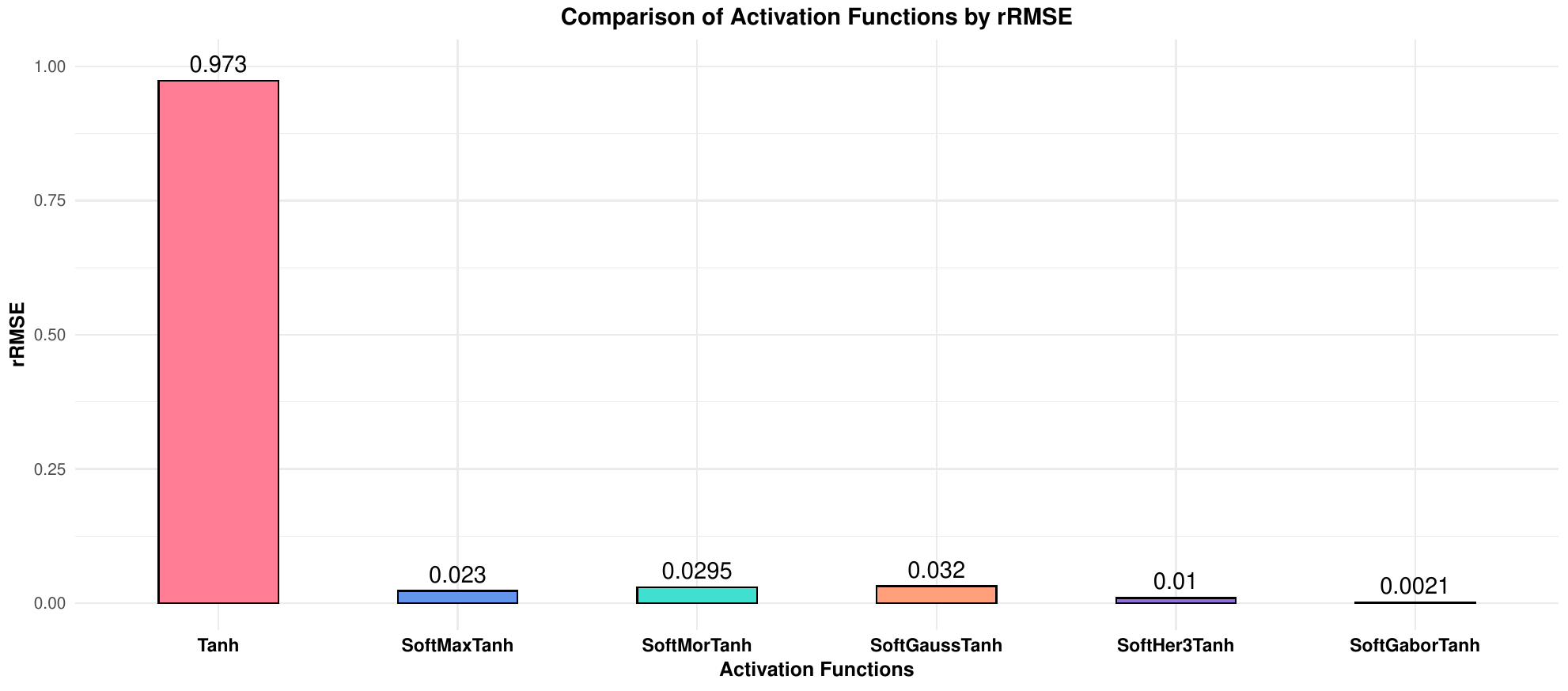}
    \caption{Relative RMSE (rRMSE) for different activation functions.}
    \label{fig:rRMSE_act}
\end{subfigure}

\caption{Performance comparison of activation functions using Loss, rMAE and rRMSE.}
\label{fig:activation_perf}
\end{figure}

\begin{figure}[htbp]
\centering

\begin{subfigure}[b]{0.8\textwidth}
    \centering
    \includegraphics[width=\textwidth]{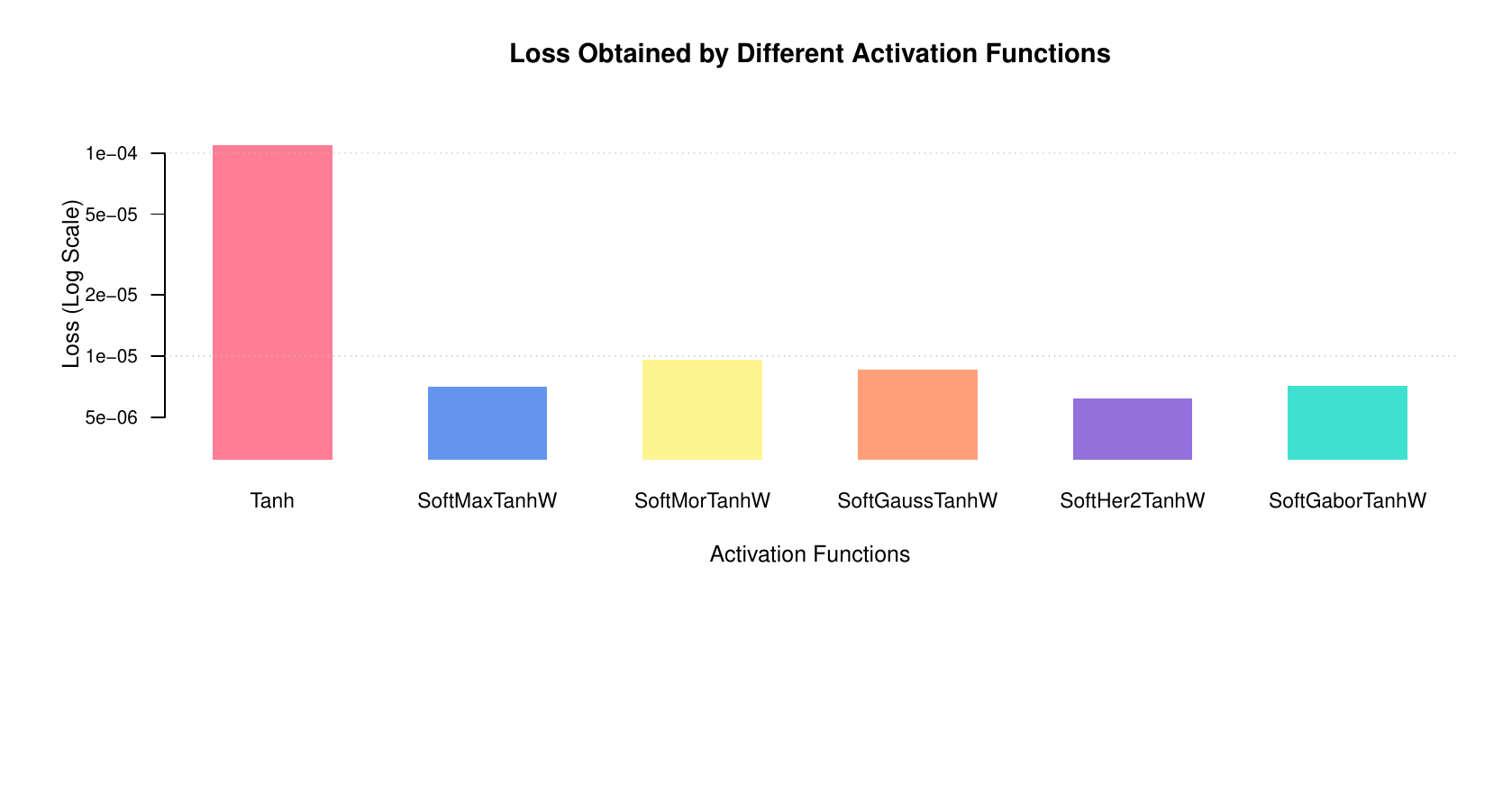}
    \caption{Loss values for different activation functions.}
    \label{fig:loss_act}
\end{subfigure}

\vspace{0.4cm}

\begin{subfigure}[b]{0.8\textwidth}
    \centering
    \includegraphics[width=\textwidth]{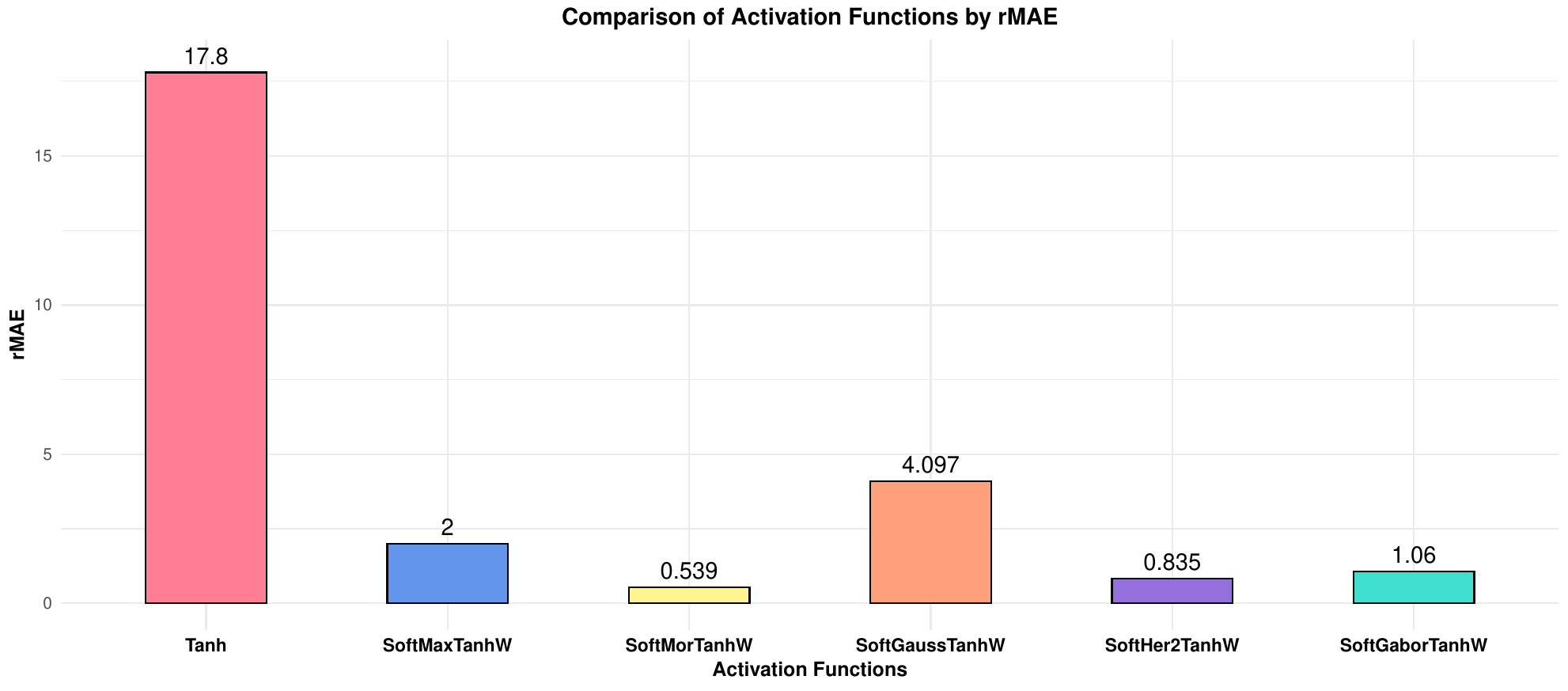}
    \caption{Relative MAE (rMAE) for different activation functions.}
    \label{fig:rMAE_act}
\end{subfigure}

\vspace{0.4cm}

\begin{subfigure}[b]{0.8\textwidth}
    \centering
    \includegraphics[width=\textwidth]{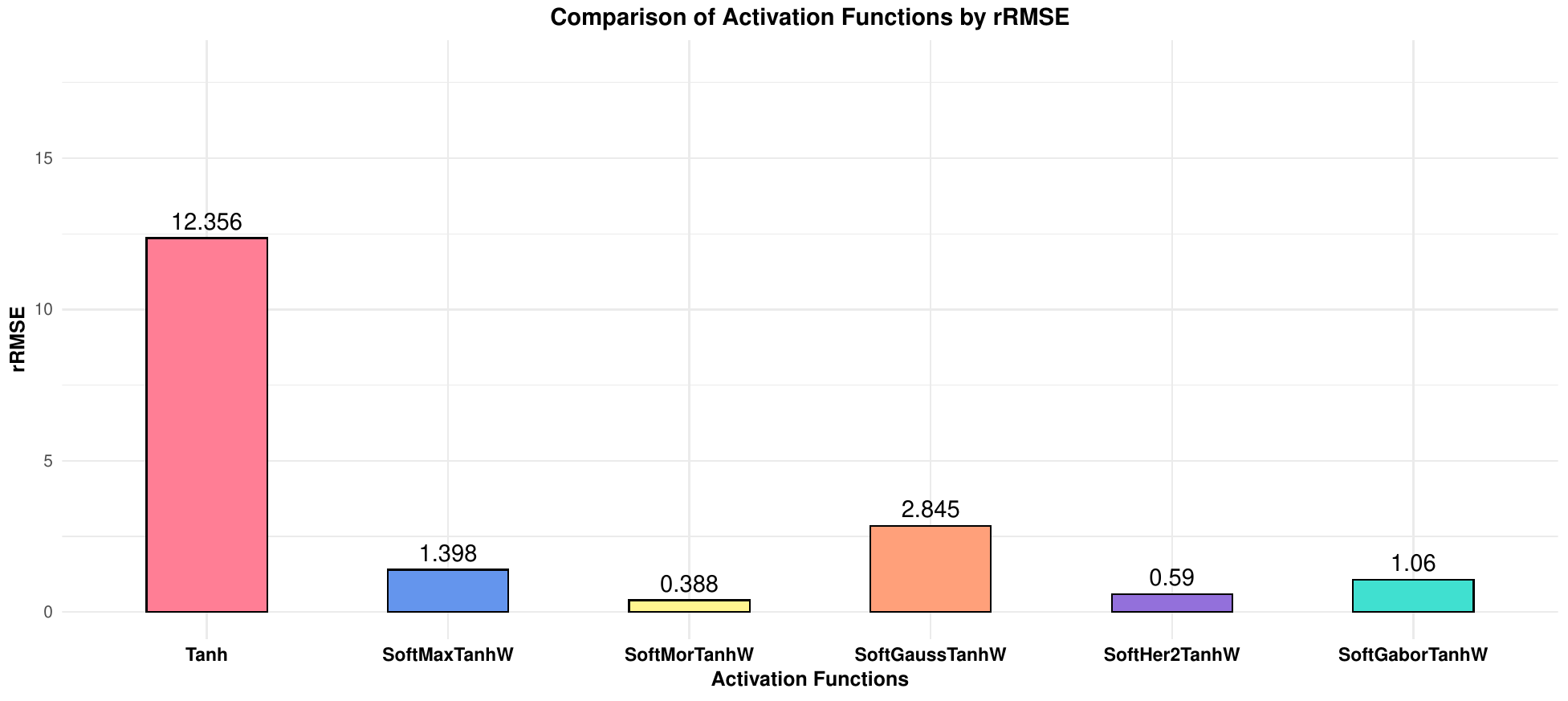}
    \caption{Relative RMSE (rRMSE) for different activation functions.}
    \label{fig:rRMSE_act_ns}
\end{subfigure}

\caption{Performance comparison of activation functions using Loss, rMAE, and rRMSE.}
\label{fig:activation_perf_last}
\end{figure}

The evaluation of the obtained data from the initial and the final experimental setting by the statistics method was performed using the RStudio program. Figures~\ref{fig:activation_perf} and~\ref{fig:activation_perf_last} represent a comparative evaluation of the activation functions.
It is clearly seen from the figures that the activation function affects the performance of the model. As one can observe from Figure~\ref{fig:loss_plot_1d_reaction}, the Loss considerably decreases when passing from one activation function to another. A very similar trend can also be observed for both \texttt{rMAE} and \texttt{rRMSE} shown in Figures~\ref{fig:rMAE_act} and~\ref{fig:rRMSE_act}, respectively. Most strikingly, except for the standard \texttt{Tanh} activation, all the proposed new activation functions produce considerably superior results for all performance metrics. In an analogous fashion, we assess the performance of the Navier–Stokes system by using the \texttt{SoftMexTanhW}, \texttt{SoftMorTanhW}, \texttt{SoftGaussTanhW}, \texttt{SoftGaborTanhW}, and \texttt{SoftHerTanhW} activation functions with the corresponding performance depicted in Figure~\ref{fig:activation_perf_last}. These figures show that the newly proposed activations overcome the deficiencies of the conventional choices.

\section{Conclusion}\label{sec:7}
The work presented here addresses deficiencies of traditional PINNs by exploiting the performance potential of wavelet-based activation functions tailored to overcome limitations of conventional activation functions, such as the commonly used \texttt{Tanh}. To validate the performance of the purpose work, four fundamental classes of PDEs were considered—the reaction equation, wave equation, convection equation, and the Navier–Stokes system—revealing that standard PINNs with \texttt{Tanh} are effective only in low-complexity settings and struggle to capture the underlying physical behavior as problem difficulty increases. One of the key features of the proposed activations is that their parameters are treated as learnable variables, allowing them to adapt dynamically during the optimization process. Overall, numerical experiments demonstrate that the wavelet-based activation functions achieve substantially higher prediction accuracy and effectively mitigate the failure modes associated with standard activation functions, as evidenced by the relative $\ell_1$ and $\ell_2$ error norms. Moreover, statistical performance evaluations depicted in a bar plot also clearly indicate higher efficacy and consistent performance. Furthermore, the proposed adaptive wavelet-inspired PINNs achieve higher accuracy and lower computational cost compared to several existing techniques, while maintaining computational efficiency and algorithmic simplicity. These enable them to function well for large-scale practical scientific purposes. The proposed activation functions retain their algorithmic simplicity, making them well-suited for practical and computationally intensive scientific applications. Future work will focus on extending training epochs, integrating hybrid Adam–L-BFGS optimization strategies to improve convergence, and developing faster and more robust PINN algorithms optimized for CPU-based architectures.

\subsubsection*{Statement of Data Availability} Data and code will be shared on request.

\subsubsection*{Declaration of Interests}
The author declares that there are no competing or conflicting interests.
\subsubsection*{Funding} This research did not receive any specific funding.
\subsubsection*{Acknowledgment}
The author acknowledges National Supercomputing Mission (NSM) for providing computing resources of ‘PARAM RUDRA’ at P G Senapathy Center For Computer Resources, Play Field
Ave, Indian Institute Of Technology, Chennai, Tamil Nadu 600036, which is implemented by
C-DAC and supported by the Ministry of Electronics and Information Technology (MeitY) and
Department of Science and Technology (DST), Government of India.

\bibliographystyle{abbrv}
\bibliography{sample}
\end{document}